\documentclass[nohyperref]{article}
\pdfoutput=1

\usepackage{microtype}
\usepackage{graphicx}
\usepackage{subfigure}
\usepackage{booktabs} 

\usepackage{hyperref}
\usepackage{url}
\usepackage{multirow}       
\usepackage[dvipsnames]{xcolor} 
\usepackage{soul}
\usepackage{colortbl}       
\usepackage{pgf}            
\usepackage{enumitem}       
\usepackage{wrapfig}        

\usepackage{bxcjkjatype}    

\usepackage[accepted]{arxiv}

\usepackage{amsmath}
\usepackage{amssymb}
\usepackage{mathtools}
\usepackage{amsthm}
\usepackage{xspace}  
\usepackage{listings} 

\usepackage[capitalize,noabbrev]{cleveref}

\theoremstyle{plain}
\newtheorem{theorem}{Theorem}[section]
\newtheorem{example}[theorem]{Example}

\theoremstyle{definition}

\theoremstyle{remark}

\definecolor{darkmagenta}{rgb}{0.55, 0.0, 0.55}
\definecolor{darkviolet}{rgb}{0.58, 0.0, 0.83}

\usepackage[textsize=tiny]{todonotes}

\definecolor{codegreen}{rgb}{0,0.6,0}
\definecolor{codegray}{rgb}{0.5,0.5,0.5}
\definecolor{codepurple}{rgb}{0.58,0,0.82}
\definecolor{backcolour}{rgb}{0.95,0.95,0.92}
\lstdefinestyle{mystyle}{
    backgroundcolor=\color{backcolour},   
    commentstyle=\color{codegreen},
    keywordstyle=\color{magenta},
    numberstyle=\tiny\color{codegray},
    stringstyle=\color{codepurple},
    basicstyle=\ttfamily\footnotesize,
    breakatwhitespace=false,         
    breaklines=true,                 
    captionpos=b,                    
    keepspaces=true,                 
    numbers=left,                    
    numbersep=5pt,                  
    showspaces=false,                
    showstringspaces=false,
    showtabs=false,                  
    tabsize=2
}
\icmltitlerunning{Interactive-Chain-Prompting: Ambiguity Resolution for Crosslingual Conditional Generation with Interaction}

\newcommand{\icp}{\textsl{\textsc{InterCPt}}\xspace}     
\newcommand{\ambigmt}{\textsl{\textsc{AmbigMT}}\xspace}  
\newcommand{\bleu}{\textsl{\textsc{bleu}}\xspace}        %
\newcommand{\bleurt}{\textsl{\textsc{bleurt}}\xspace}    %

\begin{document}

\twocolumn[
\icmltitle{Interactive-Chain-Prompting: Ambiguity Resolution for Crosslingual Conditional Generation with Interaction}



\icmlsetsymbol{equal}{*}

\begin{icmlauthorlist}
\icmlauthor{Jonathan Pilault}{comp1,mila,polymtl,equal}
\icmlauthor{Xavier Garcia}{comp1}
\icmlauthor{Arthur Bražinskas}{comp2}
\icmlauthor{Orhan Firat}{comp1}
\end{icmlauthorlist}

\icmlaffiliation{mila}{Mila}
\icmlaffiliation{polymtl}{Polytechnique Montreal}
\icmlaffiliation{comp1}{Google Research, Brain Team}
\icmlaffiliation{comp2}{Google Research, XGen Team}

\icmlcorrespondingauthor{Jonathan Pilault}{pilaultj@mila.quebec}
\icmlcorrespondingauthor{Xavier Garcia}{xgarcia@google.com}
\icmlcorrespondingauthor{Arthur Bražinskas}{abrazinskas@google.com}
\icmlcorrespondingauthor{Orhan Firat}{orhanf@google.com}

\icmlkeywords{Machine Learning, ICML}

\vskip 0.3in
]



\printAffiliationsAndNotice{\icmlEqualContribution} 

\begin{abstract}

Crosslingual conditional generation (e.g., machine translation) has long enjoyed the benefits of scaling.
Nonetheless, there are still issues that scale alone may not overcome.
A source query in one language, for instance, may yield several translation options in another language without any extra context.
Only one translation could be acceptable however, depending on the translator's preferences and goals.
Choosing the incorrect option might significantly affect translation usefulness and quality.
We propose a novel method \emph{interactive-chain prompting} --- a series of question, answering and generation intermediate steps between a \emph{Translator} model and a \emph{User} model --- that reduces translations into a list of subproblems addressing ambiguities and then resolving such subproblems before producing the final text to be translated.
To check ambiguity resolution capabilities and evaluate translation quality, we create a dataset exhibiting different linguistic phenomena which leads to ambiguities at inference for four languages. 
To encourage further exploration in this direction, we \href{https://github.com/jpilaul/interactive_chain_prompting}{\color{WildStrawberry}{\textbf{release}}} all datasets.
We note that \emph{interactive-chain prompting}, using eight interactions as exemplars, consistently surpasses prompt-based methods with direct access to background information to resolve ambiguities.

\end{abstract}




\section{Introduction}
\label{sec:intro}

\begin{figure}[ht!]
\small
\centering
\includegraphics[width=0.48\textwidth, angle=0]{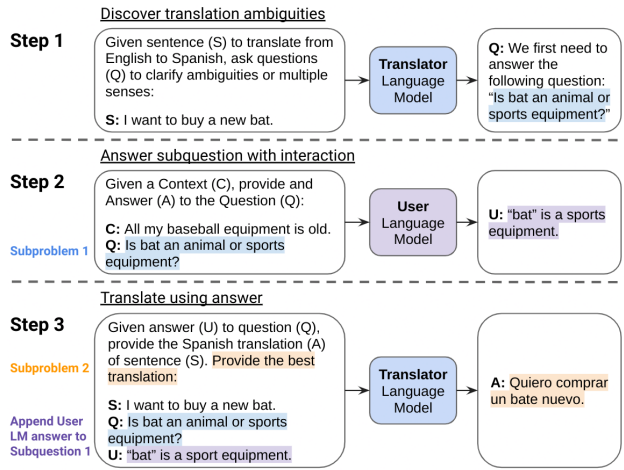}
\vspace{-0.75em}
  \caption{Interactive-Chain-Prompting (\icp).} \label{fig:interactive-steps}
\vspace{-1.5em}
\end{figure}

Transformer Language Models (LM, \citealt{NIPS2017_transformer}) pretrained on large corpora have achieved outstanding results in a variety of NLP benchmarks \cite{devlin-etal-2019-bert, NEURIPS2020_lm_few-shot}. 
Scaling the number of parameters, the size of the pretraining dataset, and the amount of computing budget gives Language Models better sample efficiency and ability to generalize for many tasks \cite{scaling_law_kalan_2020, NEURIPS2020_lm_few-shot, scaling_autogen_2020, scaling_transfer_2021_hernandez, lepikhin2021gshard, wei2022emergent}. 
However, for tasks such as commonsense and symbolic reasoning, where the solution requires multistep computation, or crosslingual conditional generation such as Neural Machine Translation (NMT), where there could be more than one plausible prediction for a given source sequence, scale alone may not be sufficient to achieve high accuracy \cite{scaling_lm_analysis_2021,ghorbani2022scaling}.

Chain-of-thought \cite{wei2022chain} and least-to-most \cite{least2most_prompt_2022} methods have demonstrated, by prompting a (large-)LM such as PaLM \cite{palm_chowdhery_2022}, that breaking down a task into subproblems that are solved sequentially greatly improves the quality of the final prediction.
Such methods demonstrate that producing intermediate sub-results that address specific aspects of a bigger problem significantly improves performance on tasks like arithmetic, math word problems, and symbolic manipulation.
While studies have investigated the translation capabilities of PaLM with various prompting strategies \cite{POMP_vilar_2022,biaoLatestPrompting}, prompting large and general purpose LMs such as PaLM to identify and solve subproblems in crosslingual conditional generation tasks such as NMT has not yet been fully explored.

\begin{figure}[ht!]
\small
\centering
\includegraphics[width=0.48\textwidth, angle=0]{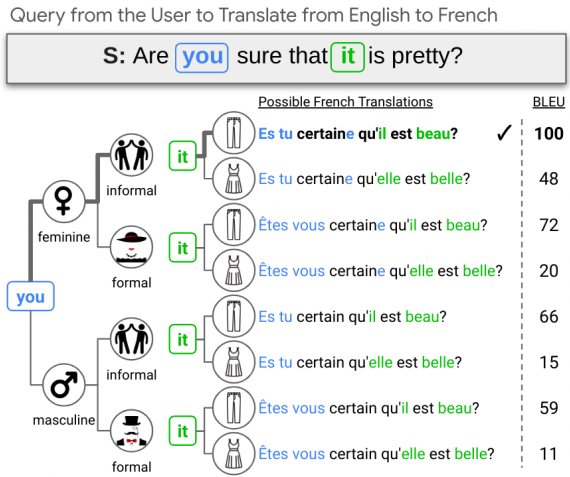}
\vspace{-0.75em}
  \caption{Translation queries with multiple possible predictions. Correctly solving subproblems around ambiguities with \textbf{\textcolor{RoyalBlue}{you}} and \textbf{\textcolor{Green}{it}} greatly affects the \bleu \protect\cite{papineni-etal-2002-bleu} translation metric.} \label{fig:ambiguity}
\end{figure}

Our approach, \emph{Interactive-Chain-Prompting} (\icp), sequentially solves translation subproblems before generating a final translation prediction.
As shown in Figure~\ref{fig:interactive-steps}, we first detect ambiguities in translation queries, then we resolve these ambiguities via question-answer interactions, and finally we generate translations.
\icp departs from other prompt-based techniques that sequentially solve subproblems in two fundamental ways:
(1) the subproblems are related but considerably different to the main task and (2) the solutions to subproblems requires interaction with another LLM.
In this paper, we will look at how intermediate computation steps and interaction might assist overcome a typical problem in automated systems when a user's ambiguous query leads to a large number of viable and potentially inaccurate answers.
In translation, for example, selecting the incorrect prediction has a significant impact on translation quality as illustrated in Fig. \ref{fig:ambiguity}.

\icp has several advantages.
First, the LM is able to identify and ask questions about translation query ambiguities with only a few in-context exemplars and no finetuning.
This is crucial since large corpora with specific target ambiguities, labels to classify each ambiguity subtypes (i.e. feminine/masculine for gender or formal/informal for formality) and context are not common and are typically low-resource.
Then, without readily available context, we rely on the \emph{User} to disambiguate translation queries.
In the absence of additional background information or context, there are limited options to solve ambiguities.
Interaction with the \emph{User} stands as a logical way to collect clarifying information.
This interaction also benefits from multiple computation steps where ambiguity resolution leads to a more precise final prediction.
Finally, the question-answer-translation interaction improves transparency and makes it easier to debug translation systems since we can assess the reasoning chain that led to an error \cite{aichain_wu2022}.
For NMT, there are two main questions to consider to make the most of out of intermediate computation steps:

\textbf{A) What subproblem are we trying to solve?}
Multistep reasoning tasks can often be explicitly decomposed into subproblems: ambiguity detection, disambiguation via Q\&A and translation.
For NMT, decomposing the translation task is not trivial. 
We assume in this work that our subproblems are ambiguities which arise when translating.
As seen in Fig.~\ref{fig:interactive-steps}, the first step in \icp is to discover and resolve the translation ambiguity subproblem.
We study five types of ambiguities: polysemous words, pronoun resolution, formality, gender-neutral names and neutral professions. 
Since datasets that cover multiple translation ambiguities and language pairs while providing context are rare, we create our own datasets (see Table~\ref{tab:other_data} in Section~\ref{app:more_datasets} for an overview of other publicly available datasets). 


\textbf{B) Where do answers to subquestions come from?}
When we apply least-to-most prompting to math word problems for example, the answers to subquestions can often be derived from the problem's text.
It is not necessarily the case for NMT where the query may not contain enough context to resolve ambiguities. 
As seen in Fig.~\ref{fig:ambiguity}, English sentence `S' does not contain enough information about ``you'' and ``it''.
The incorrect prediction made by a model leads to large variations in translation quality scores. 
With more context, the model may have the necessary information to narrow down possible predictions. 
However, in industrial applications, translation queries are often too short \cite{MT_eBay2016} or additional context is not existent.
In this work, we automate interaction between a \emph{PaLM Translator} model, that detects ambiguities, asks clarifying questions and translates, and a \emph{PaLM User} model, that has access to context and answers questions.
Both models engage in a multiturn dialog to zero-in on a narrower set of predictions. 
We argue that a type of question-answer interaction with a ``user'' is necessary to resolve ambiguous queries, especially when a user
(1) is unfamiliar with the main task and may not possess the skills to choose from many model prediction options;
(2) knows how to answer simple pointed questions about a query but may not be able or willing to decide and add appropriate context on the fly.

This work marks Large-LM's potential to learn, with a few in-context examples, how to use natural language answers to deliver results closer to a user's intent.
Our contributions are the following:
\begin{enumerate}[noitemsep,partopsep=0pt,topsep=0pt,parsep=0pt,leftmargin=4mm]

    \item We propose \icp, a new way to design crosslingual conditional generation systems that disambiguate queries via interaction (Section~\ref{sec:methodology}).
    \item We release \ambigmt, a new dataset with five specific types of ambiguities covering four languages (Section~\ref{sec:dataset}).
    \item We show that \icp achieves better translation performance and ambiguity resolution (Section~\ref{sec:results}) and improved generalization on zero-shot ambiguities (Section~\ref{sec:analysis}) over strong baselines.
    \item We provide analysis on interactions and evidence that \icp abilities emerge with scale (Section~\ref{sec:analysis}).
\end{enumerate}

\section{Interactive-Chain-Prompting (\icp)}
\label{sec:methodology}

When interacting with a model, a user may have some well-conceived query in mind that is inadvertently under-specified.
For example, a monolingual English speaker may be unaware that the pronoun ``you'' in a sentence can lead to formal or informal constructs in other languages and may therefore not provide additional information on the level of formality needed to adequately translate the text.

A human translator, when asked to translate queries with ``you'', may want to first probe the user's latent context about the query by asking clarifying questions.
In doing so, the human translator can use the answers to better align the translation to a User's request and context.
Our method endows language models (LMs) with the ability to generate a similar chain of interactions between a Translator LM and a User LM as seen in Fig. \ref{fig:interactive-steps}.
In real applications, it is expected that a human replaces the User LM.
\icp uses in-context exemplars to resolve ambiguities before completing the crosslingual conditional generation task that the model is originally asked to do.

It consists of a three step reasoning chain (See Fig.~\ref{fig:interactive-steps}):

\begin{enumerate}[noitemsep,partopsep=0pt,topsep=0pt,parsep=0pt,leftmargin=4mm]
    \item \textbf{The first step is for identifying ambiguities.} 
    The prompt in this step always contains the same constant exemplars, showing multiple queries to translate and questions about each query's ambiguities. 
    During inference, the \emph{Translator} LM uses the prompt to generate a pointed question that identifies the specific ambiguity.
    \item \textbf{The second step is for resolving ambiguities.} 
    The prompt in this step contains exemplars answering the question to the ambiguity subproblems in step one. 
    The \emph{User} LM answers each question using additional information from the provided context. 
    In real life applications, we assume that a real user has similar background information about the text to be translated.
    \item \textbf{The third step is for translating.} 
    Generated questions and answers are appended to the prompt in step 1 before the final translation is produced. 
    Constant prompts in this step demonstrate how to translate in the specified target language using only details provided by the \emph{User} LM and no-context. 
    During inference, the \emph{Translator} LM uses the prompt to generate the translation.
\end{enumerate}

\begin{table}[h]
\caption{\label{tab:data_eg} \ambigmt data examples for each ambiguity for target language \textit{x}. $\Delta$ B is the \bleu performance drop from 100 if the highlighted ambiguity is resolved incorrectly.}
\vskip 0.1in
\setlength\tabcolsep{2pt}
\scriptsize
\begin{tabular}{ p{1.0cm} | p{1.6cm} | p{2.6cm} | p{1.7cm} | r}
\toprule
\textbf{Dataset} &  \textbf{\textit{en} Query}  & \textbf{Context}   & \textbf{\textit{x} Target} & $\Delta$ B
\\\midrule
\textbf{``it'' resolution}
& He has read \textbf{\textcolor{Green}{it}} to me so many times that I've learnt \textbf{\textcolor{Green}{it}} by heart.
& - I remember when the \textbf{\textcolor{Green}{postcard}} came, Ernesto was so pleased. - He said: "Look what my Rosetta has written to me".       
& Me \textbf{\textcolor{Green}{la}} sé de memoria de tanto \textbf{\textcolor{Green}{leerla}}.
& -44
\\\hline
\textbf{Polysemy} 
& \textbf{\textcolor{orange}{head}}       
& If you don't feel well, \textbf{\textcolor{orange}{head}} home.                        
& \textbf{\textcolor{orange}{\begin{uCJK}\UTF{5148}\end{uCJK}}}
& -100
\\\hline
\textbf{Formality}
& The closer \textbf{\textcolor{RoyalBlue}{you}} can get to him, the better.        
& - I'm aware of the risks, \textbf{\textcolor{RoyalBlue}{Master Jedi}}, but I know \textbf{\textcolor{RoyalBlue}{you}} can regain Clovis' trust.
& Plus \textbf{\textcolor{RoyalBlue}{vous serez}} proche de lui, mieux \textbf{\textcolor{RoyalBlue}{cela}} sera.
& -58
\\\hline
\textbf{Gender neutral names}
& \textbf{\textcolor{darkviolet}{Blair}} should be wrapping up \textbf{\textcolor{darkviolet}{[pr]}} breakfast with Beatrice.
& - I have \textbf{\textcolor{darkviolet}{her}} doorman on retainer. - There's a fine line between surveillance and stalking. 
& Blair sollte \textbf{\textcolor{darkviolet}{ihr}} Frühstück mit Beatrice haben.
& -40
\\\hline
\textbf{Neutral professions}
& \textbf{\textcolor{blue}{[pr]}} worked previously as a businesswoman, accountant, and bank executive.
& \textbf{\textcolor{blue}{Margaret}} Mhango Mwanakatwe is a Zambian politician [...]. \textbf{\textcolor{blue}{She}}  was the director for business development [...]
& Previamente, trabajó como  \textbf{\textcolor{blue}{empresaria}}, \textbf{\textcolor{blue}{contadora}} y \textbf{\textcolor{blue}{ejecutiva}} bancaria.
& -70
\\
\bottomrule
\end{tabular}
\end{table}

\section{Ambiguity MT Datasets (\ambigmt)}
\label{sec:dataset}

In this section, we introduce \ambigmt, a dataset that covers four language pairs, for translations from English into French (en-fr), German (en-de), Spanish (en-es) or Japanese (en-ja) --- 18 sub-tasks in total.
The code and datasets are released \href{https://github.com/jpilaul/interactive_chain_prompting}{\color{WildStrawberry}{\textbf{here}}}.
The parallel translation corpora contain five types of ambiguities: ``it'' resolution, formality, polysemy, gender\footnote{Please note that due to the lack of large translation corpora with various genders and the complexity in creating non-binary gender datasets, our data is limited to feminine and masculine.} neutral names, neutral professions.
Unless otherwise specified, all datasets include 1000 diverse samples for each $\{$en-fr, en-de, en-es, en-ja$\}$ language pair extracted from Opensubtitles corpora~\cite{lison-tiedemann-2016-opensubtitles2016}. 
In Section~\ref{app:more_datasets} of the Appendix, we provide more details on datasets and describe the heuristics to identify ambiguities in each language.

\vspace{-0.6em}
\paragraph{``it'' resolution} data contains English sentences where the pronoun ``it'' does not clearly refer to a noun within the query.
In English, the pronoun ``it'' is a singular, neuter and impersonal pronoun. 
In other languages, ``it'' may translate into gender specific pronouns (either feminine or masculine) or get dropped entirely from the sentence. 
The choice depends on what the pronoun refers to.
To correctly translate, the model must first determine what ``it'' is.
In the first example of table~\ref{tab:data_eg} where the target language \textit{x} is Spanish, knowing that ``it'' is a postcard, or \emph{una tarjeta postal} in Spanish, disambiguates gender in the translation.
While the gender affects two words in the target sentence, the wrong gender choice is not only qualitatively inappropriate but also decreases quality metrics (44 \bleu score drop from 100).

\vspace{-0.6em}
\paragraph{Polysemy} is a dataset that contains words that have multiple meanings and the query is insufficiently informative to zero-in on a specific sense. 
The context uses the word within a sentence to provide the necessary background information. 
In the second example of Table~\ref{tab:data_eg} where the target language \textit{x} is Japanese, the context shows that ``head'' is a verb. 
In conjunction with the noun ``home'', we disambiguate ``head'' as ``to move in the direction of''. 
In the absence of such context, ``head'' has various senses such as ``upper part of the body'', ``side of a coin'', ``end of a hammer or tool'',  ``a toilet on a boat'', ``to hit the ball with the head'', ``to lead''.

\vspace{-0.6em}
\paragraph{Formality} is a dataset where English queries contain the pronoun ``you''.
In the target languages studied, ``you'' can be formal or informal.
As seen in the third example of table~\ref{tab:data_eg} where the target language \textit{x} is French, the speaker addresses the listener ``you'' as ``Master Jedi'' in the context, a title implying a formal style of politeness.
The formality is ambiguous without the context and may impact the generated translation quality.
Indeed, an incorrect choice in formality level changes ``vous serez'' to ``tu seras'' and ``cela'' to ``ça'', decreasing \bleu scores by 58 points from 100.

\vspace{-0.6em}
\paragraph{Gender Neutral Names} data includes queries where the name is gender neutral and ambiguous.
The fourth example in table~\ref{tab:data_eg} shows a query where the name ``Blair'' is gender neutral.
In this dataset, we replace gendered pronouns in the English query by the token \emph{[pr]} to remove hints about gender type.
From the context, the speaker employs ``her'' and we can infer that a feminine pronoun ``ihr'' should be used in the translated German text.

\vspace{-0.6em}
\paragraph{Neutral Professions} has 600 unique samples for two language pairs.
This dataset is derived from the Translated Wikipedia Biographies dataset\footnote{https://ai.googleblog.com/2021/06/a-dataset-for-studying-gender-bias-in.html} that covers $\{$en-de, en-es$\}$.
In this dataset, the gender of typically gender-neutral professional designations is not clear from the English query alone.
In the fifth example of table~\ref{tab:data_eg}, the context provides additional hints that the query is talking about ``Margeret'', also designated by the feminine pronoun ``she''.
Resolving gender allows the model to correctly translate the list of professions in the query and potentially limiting the 70 points drop in \bleu scores from 100.

\section{Related Works}
\label{sec:related_works}

\paragraph{Prompting for Cross-Lingual Generation} using Large LMs is a technique that has garnered increasing attention of late.
Works on GPT-3 \cite{NIPS2017_transformer} and PaLM \cite{palm_chowdhery_2022} show competitive $n$-shot \bleu translation results on WMT. 
The prompt demonstrations are populated with $n$ random sentence pairs taken from the WMT training corpora and evaluated on the test corpora at inference.
Orthogonal to our work, POMP \cite{POMP_vilar_2022} improves upon this PaLM-based prompting technique by explicitly optimizing for the selection of $n$ demonstration sentence pairs and obtaining results competitive with the state-of-the-art.
More recent work \cite{mtprompts_garcia_2022} using mT5 \cite{xue-etal-2021-mt5} investigated adding prompt-based natural language specifications to influence translated text properties such as formality level or dialect type.
Experiments show that prepending textual artifacts such as ``your majesty'' to the English query conditions mT5 to generate translations in a formal tone.
Our work prompts PaLM with $n$ random translation pair exemplars as well. 
Different from previous research, we prompt with exemplars to interactively discover background knowledge or clarify ambiguities before translating.

\paragraph{Interactive Machine Learning} \cite{WARE2001281,interml_fails2003,Amershi_Cakmak_Knox_Kulesza_2014} is an approach where information is interactively and iteratively supplied to a learning system.
In prior interactive translation work, machine interactivity has assisted translators in writing translations by displaying automated word suggestions that update incrementally \cite{2014-ptm,santy-etal-2019-inmt}.
The approach however is limited by drop-down menu options and requires a certain level of sophistication from the user in the \emph{target language}.
Our approach discovers preferences and background knowledge about an input query in the \emph{source language} and more flexibly adapts translations according to a user's natural language response.
The interaction is similar to Conversational AI systems where user utterances influence generated outputs.
Task or goal oriented conversational AI systems \cite{interact_qa_2013,gao-etal-2018-neural-approaches,Hussain2019ASO} are typically deployed to answer knowledge-based questions, seek information or solve basic queries (e.g. making reservations, purchase an item).
To our knowledge, our work is the first to explore conversational interaction in cross-lingual generation.

\paragraph{Resolving ambiguities} by asking for clarifications has been a recent topic of research, for QA and conversational search systems \cite{lee-etal-2019-latent,10.1145/3331184.3331265,DBLP:journals/corr/abs-2006-10174,Dhole2020ResolvingIA,clariQ_2021wang,inscit_wu2022}. 
Departing from such methods, \icp does not produce sentences from a preset list of questions but is generated from a large LM without constrain.
Concurrently to our work, \citet{krasheninnikov2022assistance} explored finetuning GPT-3 to generate clarifying questions and provide answers using human generated data from AmbigQA \cite{min-etal-2020-ambigqa} for open-domain QA.
Another GPT-3 model simulates the user and generates answers while conditioned on ground-truth clarification questions.
In contrast, our prompt-based method only needs few-shot demonstrations.
Further, our simulated user does not rely on ground-truth clarification questions to provide an answer, which could be more realistic for a number of applications (including QA, text simplication, code generation).


\section{Experimental Setup and Results}
\label{sec:results}

\begin{table*}[ht!]
\vskip -0.1in
    \footnotesize
    \centering
    \setlength\tabcolsep{5.5pt}
    \caption{ \small \label{tab:main_results}
    Translation results using an 8-shot generalist template that contains exemplars for formality, ``it'' resolution and polysemy ambiguity types. F-Acc = formality accuracy, G-Acc = gender accuracy, B@n = $\bleurt$@n. \bleu and \bleurt results for \icp labelled with $\dag$ are significantly better than all other systems based on pair-wise significance testing \cite{koehn-2004-statistical} with p = 0.05.
    }
    \vskip 0.1in
    \begin{tabular}[b]{l|l|ccc|ccc|cccc}
        \hline
        Lang. & \multirow{2}{*}{Method} & \multicolumn{3}{c|}{Formality} & \multicolumn{3}{c|}{``it'' resolution} & \multicolumn{4}{c}{Polysemy} \\
        Pairs     &                        & \textbf{\bleu} & \textbf{\bleurt} & \textbf{F-Acc.}  
                                          & \textbf{\bleu} & \textbf{\bleurt} & \textbf{G-Acc.}       
                                          & \textbf{Hit@3} & \textbf{Hit@10} & \textbf{B@3} & \textbf{B@10}     \\
                                            
        \hline
            \multirow{4}{*}{\bf en$\rightarrow$es} & \icp & \textbf{36.3}$^{\dag}$ & \textbf{77.9}$^{\dag}$  & \textbf{67}\%                                                                
                                        & \textbf{33.6}$^{\dag}$  & \textbf{78.9}$^{\dag}$ & \textbf{77}\%                                                               
                                        & \textbf{46\%} &\textbf{48\%} & \textbf{54.6}$^{\dag}$  & \textbf{56.8}$^{\dag}$                                                         \\
            & PaLM-with-context             & 34.7 & 77.1 & 64\%                                                                
                                        & 30.8 & 77.2 & 68\%                                                               
                                        & 40\% & 46\% & 46.9 & 55.1                                                         \\
            & PaLM-no-extras              & 34.6 & 77.0 & 62\%                                                                
                                        & 29.6 & 75.9 & 63\%                                                               
                                        & 33\% & 40\% & 44.9 & 51.0                                                         \\
            & Google Translate            & 31.4 & 75.3 & 50\%                                                                
                                        & 27.5 & 73.0 & 54\%                                                               
                                        & --- & --- & --- & ---                                                             \\
                                        
        \hline
            \multirow{4}{*}{\bf en$\rightarrow$fr} & \icp & \textbf{39.1}$^{\dag}$  & \textbf{70.6} & \textbf{72\%}                                                                
                                        & \textbf{35.3}$^{\dag}$  & \textbf{71.7}$^{\dag}$ & \textbf{73\%}                                                              
                                        & \textbf{46\%} & \textbf{48\%} & \textbf{46.9}$^{\dag}$  & \textbf{48.5}$^{\dag}$                                                          \\
            & PaLM-with-context         & 36.4 & 69.9 & 65\%                                                                
                                        & 33.5 & 68.4 & 68\%                                                               
                                        & 36\% & 40\% & 40.1 & 44.7                                                         \\
            & PaLM-no-extras              & 35.7 & 69.2 & 63\%                                                                
                                        & 32.3 & 66.7 & 66\%                                                               
                                        & 33\% & 37\% & 38.1 & 41.8                                                         \\
            & Google Translate            & 30.7 & 67.4 & 58\%                                                                
                                        & 29.1 & 65.4 & 61\%                                                               
                                        & --- & --- & --- & ---                                                             \\
                                        
        \hline
            \multirow{4}{*}{\bf en$\rightarrow$de} & \icp & \textbf{35.8}$^{\dag}$  & \textbf{75.0} & \textbf{69\%}                                                               
                                        & \textbf{24.0}$^{\dag}$  & \textbf{76.0} & \textbf{75\%}                                                               
                                        & \textbf{43\%} & \textbf{45\%} & \textbf{45.1}$^{\dag}$  & \textbf{47.6}$^{\dag}$                                                          \\
            & PaLM-with-context         & 33.6 & 74.6 & 61\%                                                                
                                        & 22.4 & 75.0 & 69\%                                                               
                                        & 35\% & 39\% & 36.1 & 44.9                                                         \\
            & PaLM-no-extras              & 32.5 & 74.4 & 62\%                                                                
                                        & 22.8 & 73.2 & 63\%                                                               
                                        & 32\% & 35\% & 36.7 & 41.3                                                         \\
            & Google Translate            & 27.5 & 72.3 & 53\%                                                                
                                        & 22.1 & 73.0 & 59\%                                                               
                                        & --- & --- & --- & ---                                                             \\
         \hline
            \multirow{4}{*}{\bf en$\rightarrow$ja} & \icp & \textbf{28.6}$^{\dag}$  & \textbf{69.7}$^{\dag}$ & \textbf{67\%}                                                                
                                        & \textbf{23.1}$^{\dag}$  & \textbf{72.4}$^{\dag}$ & \textbf{74\%}                                                                
                                        & \textbf{41\%} & \textbf{44\%} & \textbf{44.7}$^{\dag}$  & \textbf{47.0}$^{\dag}$                                                          \\
            & PaLM-with-context         & 26.3 & 68.0 & 60\%                                                                
                                        & 21.4 & 70.8 & 67\%                                                               
                                        & 34\% & 38\% & 35.8 & 43.8                                                          \\
            & PaLM-no-extras              & 25.9 & 67.4 & 61\%                                                               
                                        & 21.2 & 70.3 & 61\%                                                               
                                        & 30\% & 33\% & 34.6 & 37.0                                                          \\
            & Google Translate            & 23.5 & 66.7 & 50\%                                                                
                                        & 19.9 & 68.6 & 52\%                                                               
                                        & --- & --- & --- & ---                                                             \\
 
        \hline
    \end{tabular}
\vskip -0.1in
\end{table*}

In this section, we present the main cross-lingual generation results of \icp for formality, ``it'' resolution and polysemy ambiguity resolution subtasks.
We use PaLM~\cite{palm_chowdhery_2022}, a 540B-parameter decoder-only LM pretrained on primarily English-centric data with $\sim$20\% of the data obtained from non-parallel multilingual corpora.
The \emph{generalist} prompt template is composed of two formality, three polysemy and three ``it'' resolution exemplars.
All prompt-based methods are $8$-shot with the same source sentences \textit{S} to translate and corresponding translated sentences \textit{A} in the target language.
Each target language has it's own prompt template since \textit{A} differs with every language.
The simulated LM user is based on a single English-only $8$-shot prompt template for all target languages.
Example~\ref{eg:user_lm} shows the structure of an the LM user prompt exemplars for polysemy.
A complete overview of all prompts and exemplars used in experiments can be found in Sections~\ref{app:sim_template} for the User LM and Sections~\ref{app:gen_templates} for the generalist Translator LM.

\begin{example}
\label{eg:user_lm}
Given a Context (C), provide an Answer (A) to the Question (Q): \\
\textbf{S:} about\\
\textbf{C:} About 2\% of the households are enumerated using the canvasser method. \\
\textbf{Q:}  Is ``about'' an adverb that means approximately, near or a preposition that means regarding, over, surrounding? \\
\textbf{A:} ``about'' means approximately. \\
\end{example}

\paragraph{Baselines.} 
Our main baselines were chosen to compare the cross-lingual generation abilities of a large multipurpose LMs given interaction, context or no additional information.
We compare our results against two different types of prompting techniques and a commercially available multilingual baseline with the Google Cloud Translation v2 model\footnote{https://translate.google.ca/}.
Please note that we do not add other baselines since contextual NMT systems are not common and since we introduce a new dataset.
Our strongest baseline, \emph{PaLM-with-context}, is the only method that benefits from having \textbf{all of the background information required} to resolve ambiguities.
PaLM-with-context has a prompt with exemplars formulated as the one in example~\ref{eg:lm_context}.
In the example, references to \textbf{\textcolor{RoyalBlue}{you}} and \textbf{\textcolor{Green}{it}} are directly accessible in context \textit{C}.

\begin{example}
\label{eg:lm_context}
Given context (C), Translate (S) from English to French:\\
\textbf{S:} Are \textbf{\textcolor{RoyalBlue}{you}} sure that \textbf{\textcolor{Green}{it}} is pretty?\\
\textbf{C:} She was trying on a new \textbf{\textcolor{Green}{hat}}. Looking at herself in the mirror, she asked her \textbf{\textcolor{RoyalBlue}{friend}} \textbf{\textcolor{RoyalBlue}{Isabelle}}. \\
\textbf{A:} \textbf{\textcolor{RoyalBlue}{Es-tu}} certain\textbf{\textcolor{RoyalBlue}{e}} qu'\textbf{\textcolor{Green}{il}} est \textbf{\textcolor{Green}{beau}}? \\
\end{example}
\vspace{-1.1em}

To evaluate the impact of context or interaction, we also run \emph{PaLM-no-extras}, prompting without any additional information.
The structure of a PaLM-no-extras exemplar is simlar to example~\ref{eg:lm_context} without the context \textit{C}.
The model must translate the source sentence \textit{S} in the target language without knowing details about ``it'' or the level of formality to employ for ``you''.
The baseline is not only of interest for performance comparison and to evaluate model bias but also it can provide insights on the usefulness of additional background information to disambiguate queries.
Finally, we test our datasets with a multilingual and general purpose Neural Translation Model using the Google Translate API.
This baseline allows us to set performance expectations that our PaLM-no-extras model should reach.


\paragraph{Metrics.} 
Our evaluation includes the standard \bleu and \bleurt \cite{sellam-etal-2020-bleurt} automatic translation quality metrics as well as additional measures that assess specific ambiguity resolution capabilities.
For formality, we use a rule-based classifier to quantify generated sentence formality levels (F-Acc) in the target language.
We discuss details of the heuristics in Appendix~\ref{app:classifier}.
Note that the formality classifier is based on the formality data creation scripts that allowed us to automatically identify formal and informal sentences in the source corpus.
For ``it resolution'', we found that the PaLM 62B-parameter model was surprisingly accurate at identifying translated sentence genders (G-Acc).
As seen in Table~\ref{tab:gender_classification} of Appendix~\ref{app:classifier}, PaLM 62B achieves 97\% and 93\% accuracy in classifying samples of generated translations for Spanish and French respectively.
For polysemy, we found that exact match metrics did not fully describe the performance of models.
Whenever the model generated a synonym of the ground truth, the exact match metric would not consider the prediction correct.
The PaLM-no-extras polysemy exemplars are a comma-separated list of synonyms.
Our hit@$n$ measures whether the ground truth exists in the first $n$ generated words.
For example, if the model outputs the list of Spanish words [``aproximadamente'', ``cerca de'', ``alrededor de'', ``casi'', ``más o menos''], for $n=3$, hit@$3$ would return a match for a ground truth target ``cerca de'' and no-match for a ground truth target ``casi''.
To supplement the hit@$n$ metric, we also report results of a new metric that we call $\bleurt$@$n$ (B@$n$) which returns the highest \bleurt score of the first $n$ generated word phrases.
Since \bleurt captures the non-trivial semantic similarities between words using its contextual representations from BERT, we found that the metric better measures if correct synonyms were generated by the model.
Note that we did not report the Google Translate hit@$n$ or B@$n$ numbers since the API only provides single word outputs.


\paragraph{Discussion.}
Our test results for en-es, en-fr, en-de and en-ja are summarized in Table~\ref{tab:main_results}.
We first notice that \icp surpasses all other baselines.
Surprisingly, PaLM-with-context, even with all the necessary background to resolve ambiguities, significantly lags behind \icp on F-Acc. for formality, G-Acc. for ``it resolution'' and both hit@$3$ and B@$3$ for polysemy.
This results suggests that the multistep computation approach of fist resolving the ambiguity subproblems and then generating text has an advantage over other baselines.
\bleu scores are also 2-3 points higher while \bleurt scores are only slightly higher.
This suggest that \icp generates sentences syntactically much closer to the ground truth while conserving the correct semantics.

\section{Analysis}
\label{sec:analysis}

\begin{table}[ht!]
\vskip -0.1in
    \footnotesize
    \centering
    \setlength\tabcolsep{5pt}
    \caption{ \small \label{tab:generalization}
    Translation results on unseen ambiguity subproblems using the Gender Neutral Names data and with added unseen domain using the Neutral Professions data. \icp results labelled with $\dag$ are significantly better with p = 0.05.
    }
    \vskip 0.1in
    \begin{tabular}[b]{l|l|ccc}
        \hline
        Pair & Method & \textbf{\bleu} & \textbf{\bleurt} & \textbf{G-Acc.}     \\
        \hline
        \rowcolor{blue!20}\multicolumn{5}{c}{Gender Neutral Names --- unseen ambiguities}                                                          \\
            \multirow{4}{*}{\bf en$\rightarrow$es} & \icp & \textbf{31.8}$^\dag$ & \textbf{74.1}$^\dag$ & \textbf{76\%}                                                                
                                        \\
            & PaLM-with-context             & 29.9 & 72.4 & 66\%                                                                
                                        \\
            & PaLM-no-extras              & 30.9 & 71.6 & 59\%                                                                
                                        \\
            & Google Translate            & 27.8 & 66.1 & 56\%                                                                
                                        \\
                                        
        \hline
            \multirow{4}{*}{\bf en$\rightarrow$fr} & \icp& \textbf{31.0} & \textbf{63.5}$^\dag$ & \textbf{71\%}                                                                
                                        \\
            & PaLM-with-context             & 29.5 & 62.6 & 64\%                                                                
                                        \\
            & PaLM-no-extras              & 30.0 & 60.9 & 63\%                                                                
                                        \\
            & Google Translate            & 24.5 & 57.7 & 56\%                                                                
                                        \\
                                        
        \hline
            \multirow{4}{*}{\bf en$\rightarrow$de} & \icp & \textbf{17.9}$^\dag$ & \textbf{72.2} & \textbf{73\%}                                                                
                                        \\
            & PaLM-with-context             & 15.6 & 71.5 & 67\%                                                                
                                        \\
            & PaLM-no-extras              & 15.2 & 70.8 & 61\%                                                                
                                        \\
            & Google Translate            & 17.1 & 67.1 & 55\%                                                                
                                        \\
                                        
        \hline
            \multirow{4}{*}{\bf en$\rightarrow$ja} & \icp &  \textbf{16.1}$^\dag$ & \textbf{70.3}$^\dag$ & \textbf{71\%}                                                                
                                        \\
            & PaLM-with-context             & 14.7 & 69.1 & 65\%                                                                
                                        \\
            & PaLM-no-extras              & 14.4 & 68.3 & 60\%                                                                
                                        \\
            & Google Translate            & 14.1 & 66.0 & 54\%                                                                
                                        \\
        \rowcolor{blue!20}\multicolumn{5}{c}{Neutral Professions --- unseen ambiguities + unseen domain}                                            \\
            \multirow{4}{*}{\bf en$\rightarrow$es} & \icp & \textbf{37.3} & 75.8 & \textbf{70\%}                                              
                                        \\
            & PaLM-with-context             & 37.1 & \textbf{76.1} & 69\%                                                       
                                        \\
            & PaLM-no-extras              & 35.5 & 75.7 & 59\%                                                                
                                        \\
            & Google Translate            & 37.0 & 72.7 & 56\%                                                                
                                        \\
                                        
        \hline
            \multirow{4}{*}{\bf en$\rightarrow$de} & \icp & \textbf{14.3} & 70.0 & \textbf{68\%}                                              
                                        \\
            & PaLM-with-context             & 14.0 & \textbf{71.9} & 66\%                                                       
                                        \\
            & PaLM-no-extras              & 12.2 & 70.0 & 62\%                                                                
                                        \\
            & Google Translate            & 13.8 & 67.2 & 54\%                                                                
                                        \\
        \hline
    \end{tabular}
\vskip -0.1in
\end{table}

In this section, we analyse interesting behaviors about our approach such as ambiguity generalization in Subsection~\ref{sec:generalization}, the importance of ambiguity resolution specialization in Subsection~\ref{sec:gen_vs_spe}, the effects of scale for both the Translator LM in Subsection~\ref{sec:translator_scale} and User LM in Subsection~\ref{sec:user_scale}, an error analysis in Subsection~\ref{sec:error_analysis} and bias in generated outputs in Subsection~\ref{sec:bias}.

\subsection{How does interaction generalize?}
\label{sec:generalization}

\begin{figure*}[ht!]
\vskip -0.1in
\centering
\includegraphics[width=0.9\textwidth, angle=0]{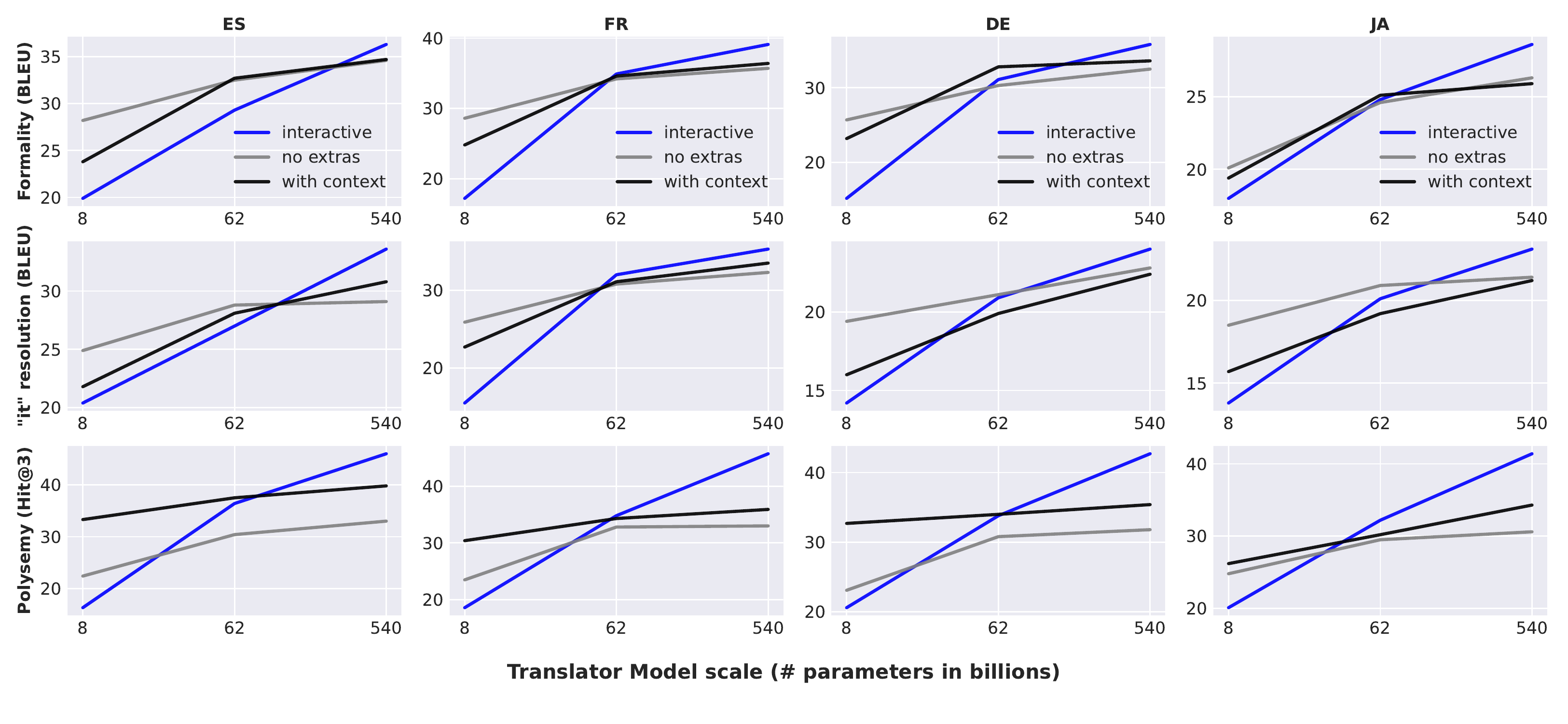}
\vskip -0.2in
  \caption{\label{fig:scale} \small \icp enables large LMs to solve ambiguity subproblems in cross-lingual generation. The multistep  disambiguate-translate capability is an emergent ability that is reached at higher parameter scales. Note that interactive = \icp.}
\end{figure*}

In Table~\ref{tab:generalization}, we provide translation test results on two held-out datasets that are described in Section~\ref{sec:dataset}: (1) Gender Neutral Names and (2) Neutral Professions.
We use the same \emph{generalist} prompt template as in Section~\ref{sec:results} with exemplars that cover only formality, ``it'' resolution and polysemy.
Specifically, our exemplars for both the Translator LM and the User LM do not contain exemplars to resolve the gender for a person's name or profession.
We observe that on the Gender Neutral Names dataset \icp performs best on \bleu and \bleurt and is much more able to resolve ambiguities with 6 to 10 points G-Acc improvements over Palm-with-context.
On the Neutral Professions data, where test samples are taken from a different domain (Wikipedia biographies instead movie scripts), Palm-with-context and \icp have similar performances.
It is possible that PaLM-with-context benefits from additional sentences in the context to better determine the style of the output.
Nonetheless, \icp provides a 1-2 point increase on G-Acc.

\subsection{Are specialist better than generalist prompts?}
\label{sec:gen_vs_spe}

\begin{figure}[ht!]
\vskip -0.1in
\centering
\includegraphics[width=0.48\textwidth, angle=0]{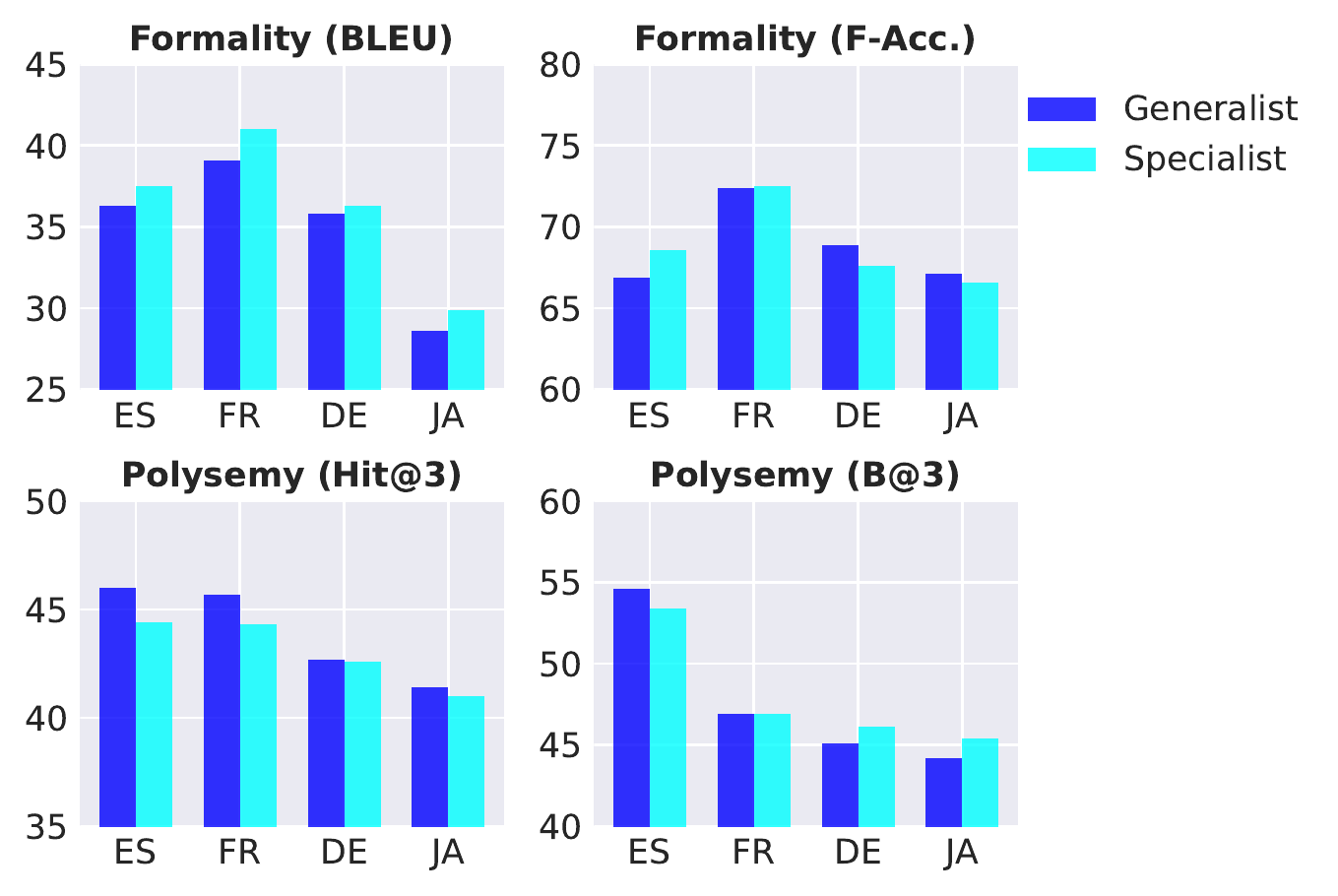}
\vskip -0.15in
\caption{
\label{fig:spe_vs_gen} \small Generalist vs Specialist prompt templates for Spanish (ES), French (FR), German (DE) and Japanese (JA) targets.}
\vskip -0.1in
\end{figure}

So far, we have studied a \emph{generalist} $8$-shot template covering three different types of ambiguities with at most three exemplars per ambiguity.
In Fig.~\ref{fig:spe_vs_gen}, we present results of \emph{specialist} template that only covers one type ambiguity at the time (either all formality or all polysemy).
Interestingly, specialization does not seem to provide much additional benefit in resolving ambiguities as evidenced by F-Acc, Hit@$3$ and B@$3$ results that are on par and often lower than the \emph{generalist} approach.
However, the \emph{specialist} template does have a higher \bleu score, implying greater syntactic alignment with the target translation when more ambiguity-specific exemplars are added.

\subsection{Are interactive generation abilities emergent?}
\label{sec:translator_scale}

We show in Fig.~\ref{fig:scale} for each prompt template the effects of scaling PaLM parameters on the performance of formality, ``it'' resolution and polysemy for Spanish (ES), French (FR), German (DE) and Japanese (JA) target languages.
Please note that while we vary the parameter count (8B, 62B and 540B) of the Translator LM, the User LM is a 540B parameters PaLM model for all experiments.
The plots provide interesting insights. 
First, at the 8B parameter scale, PaLM-no-extras performs best across all languages for Formality and ``it'' resolution across all language pairs.
Neither context or interaction seem to provide benefits to translation.
Second, at the 62B parameter scale, the PaLM-with-context and \icp methods have on par performances.
Context or interaction in this case are only clearly beneficial for polysemy.
Third, the PaLM 540B parameter \icp outpaces other prompt-based methods across language pairs and ambiguity subproblems.
At this stage, baselines scaling trend decelerates, with \emph{scaling curves flattening}, compared to \icp.
It shows that \icp is an emergent ability of model scale \cite{wei2022emergent}.
We conjecture that the emergent behavior of \icp is due to a better ability to ask questions and incorporate answers before generating final prediction.

\subsection{How important is User LM parameter scale?}
\label{sec:user_scale}

\begin{figure}[ht!]
\centering
\includegraphics[width=0.45\textwidth, angle=0]{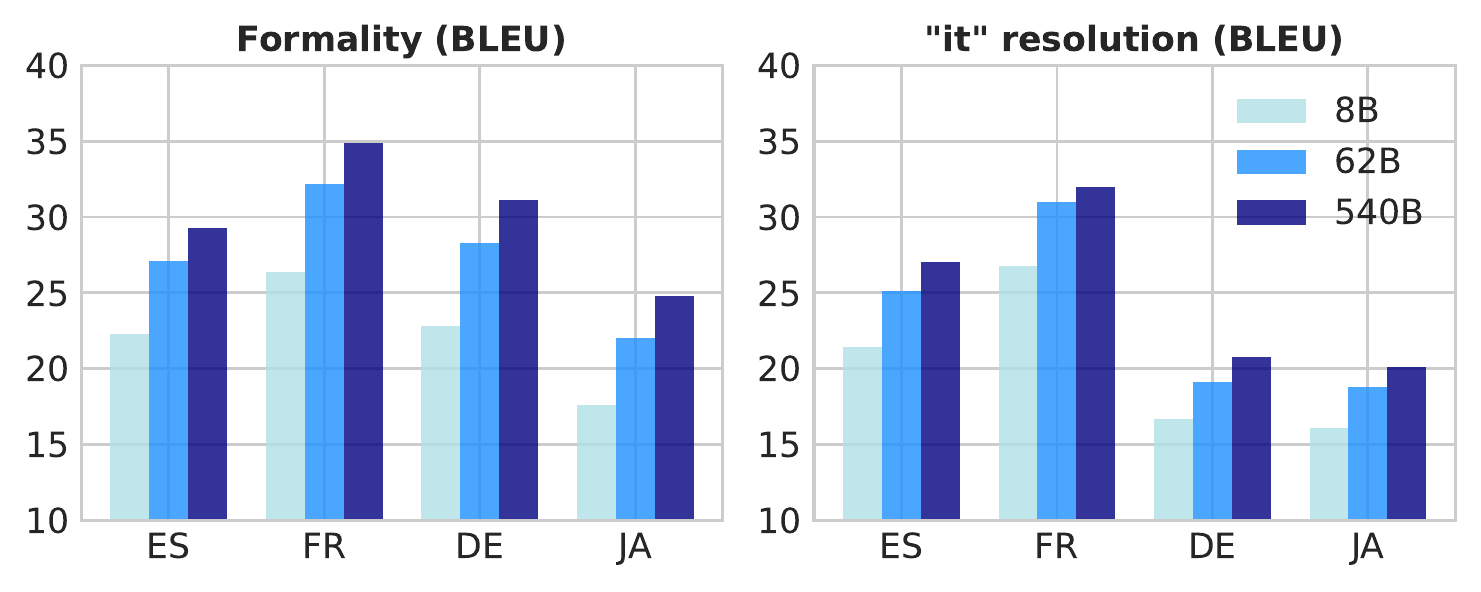}
\vskip -0.1in
  \caption{\label{tab:scale_user} \small Scaling Simulated User LM improves the performance of a 62B Translator LM model.}
\end{figure}

While the User LM allows us to automate the evaluation of interactivity for cross-lingual generation, it is not clear if the quality of the answer to the Translator LM questions impact performance.
We hypothesize that a larger User LM model would provide higher quality answers and allow the Translator LM to better generate translated text.
Fig.~\ref{tab:scale_user} shows that, when the Translator LM is a 62B PaLM model, a higher parameter User LM improve overall performance.
It is therefore possible that answer quality has a significant impact on translation quality and that human-generated answers can further improve overall performance.

\subsection{Can interaction help solve NLG bias issues?}
\label{sec:bias}

Gender bias is a common phenomenon in automated NMT systems \cite{metric_bias_borkman2019,stanovsky-etal-2019-evaluating,saunders-byrne-2020-reducing}.
Even when there are explicit gender pronouns in the input query or in the context,  NMT systems generated text tends to be masculine when translated into languages with grammatical gender \cite{stanovsky-etal-2019-evaluating,saunders-byrne-2020-reducing,stafanovics-etal-2020-mitigating,wang-etal-2022-measuring}.

To measure gender bias, all generated translations are passed through the gender classifier for the ``it'' resolution balanced dataset.
Similarly, to measure formality bias, generated translations are passed through the formality classifier for the formality balanced dataset.
NMT systems can also suffer from formality bias~\cite{rippeth-etal-2022-controlling}.
However, we notice that \icp is much closer to evenly producing masculine and feminine sentences.
Our results shows that interactive ambiguity resolution via multistep computation better addresses gender and formality biases.

\begin{figure}[ht!]
\centering
\includegraphics[width=0.48\textwidth, angle=0]{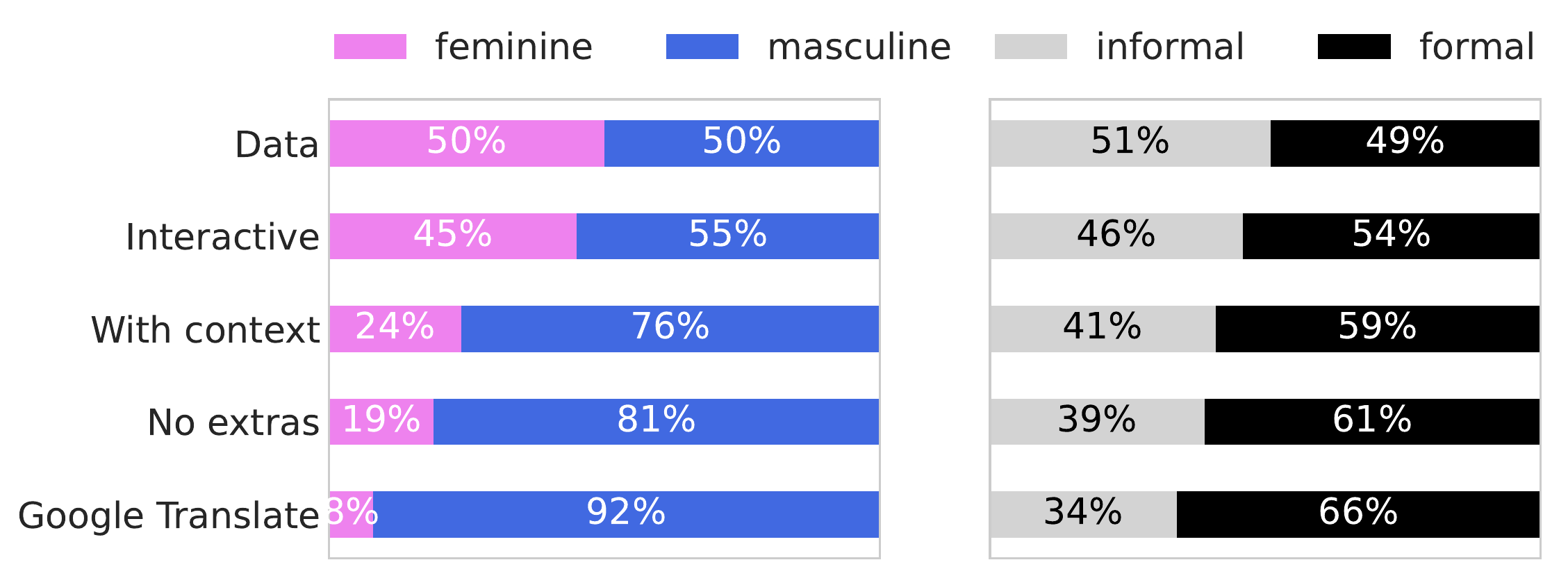}
\vskip -0.1in
  \caption{\label{tab:bias} \small Bias in generated translations for French and Spanish on ``it'' resolution (left) and formality (right).}
\end{figure}

\subsection{When is context better than interaction?}
\label{sec:error_analysis}

\begin{figure}[ht!]
\centering
\includegraphics[width=0.45\textwidth, angle=0]{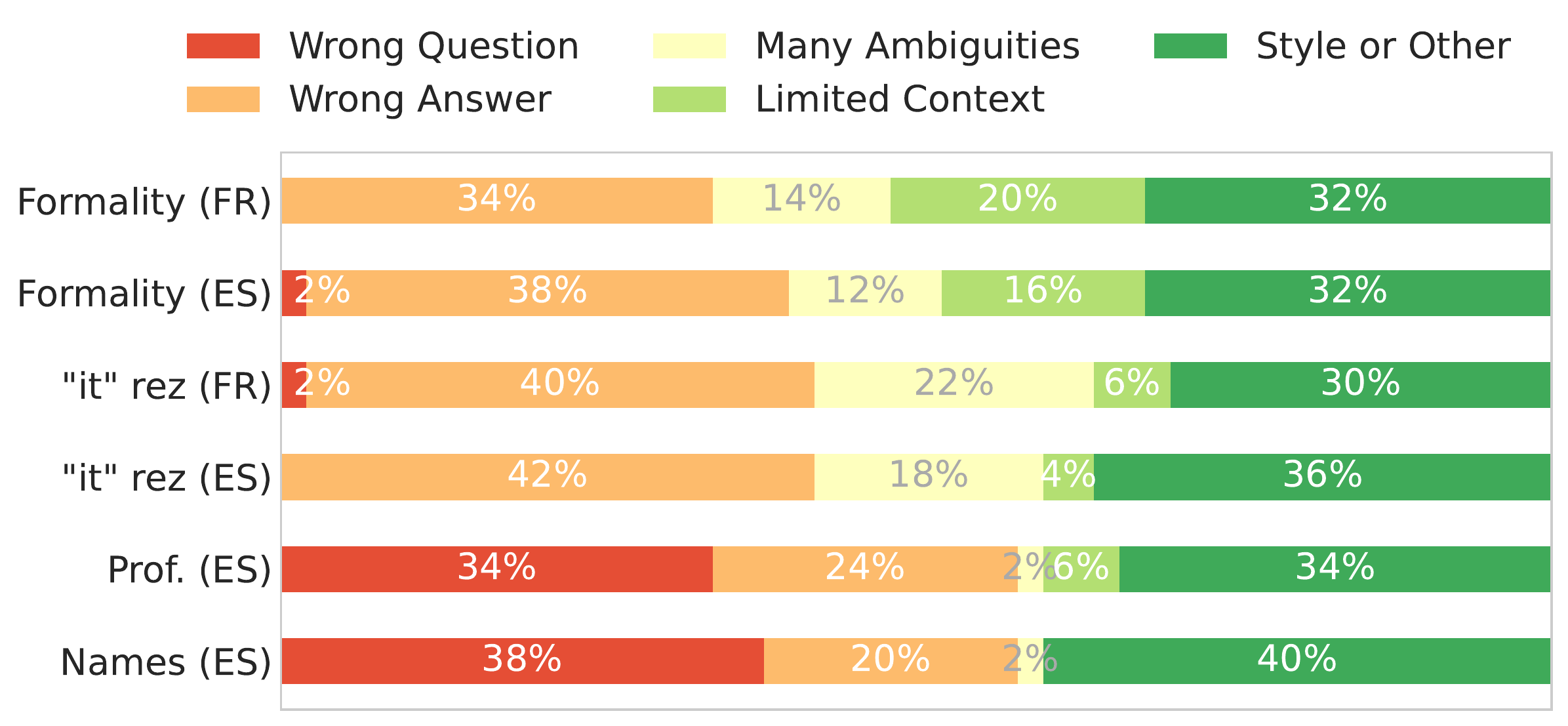}
  \caption{\label{tab:error_analysis} \small Error analysis. rez = ``it'' resolution, Prof. = Neutral profession, Names = Gender Neutral Names}
\vskip -0.1in
\end{figure}

\begin{table}[ht!]
\caption{\label{tab:error_eg} Examples of interaction chain errors.}
\vskip 0.1in
\setlength\tabcolsep{2pt}
\scriptsize
\begin{tabular}{ p{1.0cm} | p{1.8cm} | p{2.4cm} | p{2.4cm}}
\toprule

\textbf{Error Type} &  \textbf{\textit{en} Query (S) and Question (Q)}  & \textbf{Sim User Context (C) and Answer (A)}   & \textbf{Observation}
\\\midrule

\textbf{Wrong Question} 
& \textbf{S}: But I swear to you it wasn't me. \newline \textbf{Q}: What does ``it'' refer to?
& \textbf{C}: I just thought that he'd blame me for predicting his death [...]. \newline \textbf{A}: ``it'' is death
& \textbf{S} can be translated without information on ``it'' and did ask a question to disambiguate formality.
\\\hline

\textbf{Wrong Answer} 
& \textbf{S}: Develop it further, Leonard. \newline \textbf{Q}: What does ``it'' refer to?
& \textbf{C}: -Get me a complete rundown on Miller [...]. -That's a good idea. \newline \textbf{A}: ``it'' is a plan
& ``plan'' is masculine in fr and es. However, ``it'' refers to ``idea'', which is feminine in fr and es.
\\\hline

\textbf{Many Ambiguities} 
& \textbf{S}: If anyone asks, you're a relief worker. \newline \textbf{Q}: Who does ``you'' refer to? 
& \textbf{C}: -Okay, so I'm going to go with you. -White girls don't do runs. \newline \textbf{A}: 'informal' since the speaker talking to a friend ``Aaron''
& The answer is correct however the name Aaron is gender neutral and was resolved incorrectly, impacting ``worker'' translation.
\\\hline

\textbf{Limited Context} 
& \textbf{S}: I'll bring it right over. \newline \textbf{Q}: What does ``it'' refer to?
& \textbf{C}: -You didn't get it? -Really? -Just a second... \newline \textbf{A}: ``it'' is a harp
& ``harp'' is likely wrong. We cannot determine what ``it'' is from the given context.
\\\hline

\bottomrule
\end{tabular}
\end{table}

In this section, we provide analysis that describes common areas of improvement for \emph{generalist interactive-chain prompting}.
We first isolated test samples for French and Spanish for four ambiguities (formality, ``it'' resolution, neutral professions and gender neutral names) where the \bleurt scores were less than or equal to PaLM-with-context scores.
We then randomly sampled 50 interactions and manually analysed the interaction chains (query, question, context, answer, translation).
This led us to five types of errors: 
(1) wrong question, when the Translator LM asked a question not related to the ambiguity;
(2) wrong answer, when the User LM did not provide correctly disambiguate;
(3) many ambiguities, when the query had multiple unresolved ambiguities or the User LM answer also contained ambiguities;
(4) limited context, when the context was not sufficiently informative to resolve ambiguities;
(4) style or other, when generated translated text had discernible differences with the ground truth.
Fig.~\ref{tab:error_analysis} shows that the majority of errors are from wrong User LM answers for formality and ``it'' resolution.
This partially confirms our hypothesis in Subsection~\ref{sec:user_scale}.
For tasks involving unseen ambiguities, the majority of errors come from the Translator LM with 68\% to 78\% of sample chains having the wrong question or noticeable differences in generated translated text style or form.
We provide examples of interaction chains for each type of error in Table~\ref{tab:error_eg}.

\section{Conclusion}
\label{sec:conclusion}

We propose \emph{interactive-chain prompting} (\icp), a prompt-based interactive multistep computation technique that first resolves cross-lingual ambiguities in the input queries and then performs conditional text generation.
We have created and released a new datasets that covers five ambiguities: formality, ``it'' resolution, polysemy, gender neutral names and neutral professions for four different language pairs.
Empirical results show that \icp outperforms other prompt-based techniques that have access to all background information and context to directly resolve ambiguities.
We find that \icp MT is an emergent property of parameter scale that allows Large LMs to perform interactive generation tasks while other prompt-based techniques exhibit flattening scaling curves.
\icp can be considered a step forward more efficiently interacting with machine learning systems.

\section*{Acknowledgements}

For all the useful discussions and comments, we thank George Foster, Colin Cherry, Rick Genter, Patrick Fernandes and Jason Wei. 
For feedback on German and Japanese templates and translation examplars used, we thank Julia Kreutzer, Anja Austermann and Mikio Hirabayashi.

\nocite{multilingual_cot_shi2022}

\clearpage
\bibliography{main}
\bibliographystyle{icml2022}

\clearpage
\appendix
\onecolumn

\section{More details on \ambigmt ambiguity datasets}
\label{app:more_datasets}

In this section, we provide additional information on what the datasets contain and how they were created.
As mentioned in Section~\ref{sec:intro}, we did not find datasets that covered multiple ambiguities for multiple language pairs.
We provide an overview of publicly available datasets in Table~\ref{tab:other_data}.
Upon manual inspection of samples from other public datasets, we found that translation queries were often ($>$ 50\%) unambiguous since the translation query contained enough information and did not need to rely on the provided context.
We inspected 200 samples from \ambigmt and found that only $\sim$3\% of queries did not need context to disambiguate the linguistic phenomena.

\begin{table}[ht!]
\caption{
\label{tab:other_data}
Other MT datasets that contain specific linguistic phenomena and provide context. \\
en = English, de = German, fr = French, ru = Russian, zh = Mandarin Chinese, ja = Japanese.
}
\small
\centering
\begin{tabular}{| l | l | l | c |}
\toprule
\textbf{Dataset Source} &  \textbf{Language Pairs} & \textbf{Linguistic Phenomena} & \textbf{Total Test Data Size}\\

\midrule
        \citeauthor{muller-etal-2018-large}         &  en$\rightarrow$de & (1) ``it'' pronoun resolution & 12,000\\
        \hline
        \citeauthor{bawden-etal-2018-evaluating}    &  en$\rightarrow$fr & (1) Anaphora resolution, (2) lexical cohesion & 900 \\
        \hline
        \citeauthor{voita-etal-2019-good}           &  en$\rightarrow$ru & (1) Ellipsis, (2) lexical cohesion & 6,000 \\
        \hline
        \multirow{3}{*}{\citeauthor{voita-etal-2019-good}}  &  de$\rightarrow$en & \multirow{3}{*}{(1) ``it'' pronoun resolution, (2) lexical cohesion} & \multirow{3}{*}{6,090} \\
                                                            & zh$\rightarrow$en  & & \\
                                                            & en$\rightarrow$ru  & & \\
        \hline
        \multirow{4}{*}{\ambigmt (\textbf{ours})}  &  en$\rightarrow$es & \multirow{2}{*}{(1) ``it'' pronoun resolution, (2) gender neutral names} & \multirow{4}{*}{17,200} \\
                                          & en$\rightarrow$fr  & & \\
                                          & en$\rightarrow$de  & \multirow{2}{*}{(3) neutral professions, (4) polysemy, (5) formality} & \\
                                          & en$\rightarrow$ja  & & \\  
\bottomrule
\end{tabular}
\end{table}

\subsection{Dataset statistics}

We present in Table~\ref{tab:data_stats} the data statistics for \ambigmt.
For polysemy, the total senses per word is the number of different definitions or meanings found for a specific source English word.
Each ambiguity is well balanced across classes formal/informal or feminine/masculine.
The Neutral Professions dataset is derived from the Translated Wikipedia Biographies dataset\footnote{https://ai.googleblog.com/2021/06/a-dataset-for-studying-gender-bias-in.html} that only covers $\{$en-es, en-de$\}$ language pairs.

\begin{table}[ht!]
\caption{
\label{tab:data_stats}
\ambigmt data statistics of each type of class and language pair. \\
Form = formal, Inform = informal, Mas = Masculine, Fem = Feminine, res = resolution, Prof = Profession.
}

\centering
\begin{tabular}{| c | c | c | c | c | c | c | c | c | c | c |}
\toprule
\textbf{Language} & \textbf{Total} & \textbf{Polysemy} &  \multicolumn{2}{c|}{\textbf{Formality}} & \multicolumn{2}{c|}{\textbf{``it'' res.}} & \multicolumn{2}{c|}{\textbf{Neutral Names}} & \multicolumn{2}{c|}{\textbf{Neutral Prof.}} \\
\textbf{Pair}     & \textbf{Examples} & Senses/Word       &  Form.  & Inform.    & Mas. & Fem.   & Mas. & Fem.   & Mas. & Fem.   \\
\midrule
\bf en$\rightarrow$es & 4600 & 3.6 & 49\% & 51 \% & 50\% & 50\% & 51\% & 49\% & 52\% & 48\% \\
\bf en$\rightarrow$de & 4600 & 3.1 & 50\% & 50 \% & 52\% & 48\% & 50\% & 50\% & 53\% & 47\% \\
\bf en$\rightarrow$fr & 4000 & 3.3 & 49\% & 51 \% & 50\% & 50\% & 51\% & 49\% & --- & --- \\
\bf en$\rightarrow$ja & 4000 & 3.0 & 50\% & 50 \% & 52\% & 48\% & 53\% & 47\% & --- & --- \\

\bottomrule
\end{tabular}
\end{table}

\subsection{\ambigmt data creation tools, process and heuristics}
\label{app:amb_heuristics}
In this section, we present the steps, tools and heuristics used to detect ambiguities.
For polysemy, formality, ``it'' resolution, gender neutral names, we extract the data from OpenSubtitles corpora and neutral professions from Translated Wikipedia Biographies.
The source data that was used consists of parallel sentence level pairs.
We first detect a sentence that has a specific ambiguity and extract the context by taking three to five preceding English sentences, depending on sentence size.
For Polysemy, the context is an English sentence that contains the polysemous word that will be translated.
The code and datasets are released \href{https://github.com/jpilaul/interactive_chain_prompting}{\color{WildStrawberry}{\textbf{here}}}.

\subsubsection{Polysemy}

We provide the following list of steps to create the polysemy dataset for all languages:

\begin{enumerate}[noitemsep,partopsep=0pt,topsep=0pt,parsep=0pt,leftmargin=4mm]
    \item Extract polysemous words from Wordnet. \cite{miller-1994-wordnet} using the NLTK toolkit \cite{bird-loper-2004-nltk}\footnote{See example in  \url{https://www.nltk.org/howto/wsd.html}}.
    \begin{itemize}[noitemsep,topsep=3pt]
        \item Create a list of English words.
        \item Compute the number of definitions per word without counting definitions with synonym overlap.
        \item Extract polysemous words ($w_e$) with more than three definitions and a word length greater than four.
    \end{itemize}
    \item For each Polysemous English word $w_e$, extract a list $l_x = \{w_{x1}, \dotsc,w_{xN}\}$ of possible word translations using the Google Cloud Translation v2 API, where $x \in \{\text{es}, \text{fr}, \text{de}, \text{ja}\}$ is the target language.
    \item For each Polysemous English word $w_e$ and each target language $x \in \{\text{es}, \text{fr}, \text{de}, \text{ja}\}$:
    \begin{itemize}[noitemsep,topsep=3pt]
        \item Find a sentence that contains the word $w_e$ in the OpenSubtitle dataset.
        \item If the parallel sentence contains one of the translated word $w_{xi} \in l_x$ from step 2 and no other translated word, keep the English sentence as context.
   \end{itemize}
\end{enumerate}

\subsubsection{Formality}

Each language has specific formality rules. For Japanese, we direct the reader to our public code: \url{https://github.com/jpilaul/interactive_chain_prompting}. We provide the following list of steps to create the formality dataset for Spanish, French and German:

\begin{enumerate}[noitemsep,partopsep=0pt,topsep=2pt,parsep=2pt,leftmargin=8mm]
    \item Find a sentence that contains ``you'' or ``your'' and that has word count less than 20, in the English OpenSubtitle corpus.
    \item Select parallel sentences for each target language $x \in \{\text{es}, \text{fr}, \text{de}, \text{ja}\}$ that meet the following criteria.
    \item If $x == \text{es}$, check the following in parallel Spanish sentence (all checks are initialized to $\textsc{false}$):
    \begin{itemize}[noitemsep,topsep=3pt]
        \item If all verbs finish by ``s'', ``ste'' or ``os'', then is\_verb\_informal = $\textsc{true}$.
        \item If any pronouns is ``usted'', then is\_pronoun\_formal = $\textsc{true}$.
        \item If any pronouns is in [``tú'',``tu'',``te'', ``vos'', ``vosotros''], then is\_pronoun\_informal = $\textsc{true}$.
        \item If any determinants is ``su'', then is\_determinant\_formal = $\textsc{true}$.
        \item If any determinants is in [``tu'',``vosotros'', ``vosotras''] then is\_determinant\_informal = $\textsc{true}$.
        \item is\_informal = is\_verb\_informal and  is\_pronoun\_informal and is\_determinant\_informal.
        \item is\_formal = is\_pronoun\_formal and is\_determinant\_formal.
    \end{itemize}
    \item If $x == \text{fr}$, check the following in parallel French sentence (all checks are initialized to $\textsc{false}$):
    \begin{itemize}[noitemsep,topsep=3pt]
        \item If any verbs finish by ``x'', ``s'' or ``ons'', then is\_verb\_informal = $\textsc{true}$.
        \item If any verbs finish by ``ez'', then is\_verb\_formal = $\textsc{true}$.
        \item If one of the pronouns is ``vous'', then is\_pronoun\_formal = $\textsc{true}$.
        \item If one of the pronouns is ``tu'', then is\_pronoun\_informal = $\textsc{true}$.
        \item If one of the determinants is in [``vos'',``votre''], then is\_determinant\_formal = $\textsc{true}$.
        \item If one of the determinants is in [``tes'',``ton'', ``ta'', ``toi''] then is\_determinant\_informal = $\textsc{true}$.
        \item is\_informal = is\_verb\_informal and  is\_pronoun\_informal and is\_determinant\_informal.
        \item is\_formal = is\_verb\_formal and is\_pronoun\_formal and is\_determinant\_formal.
    \end{itemize}
    \item If $x == \text{de}$, check the following in parallel German sentence (all checks are initialized to $\textsc{false}$):
    \begin{itemize}[noitemsep, topsep=3pt]
        \item If ``!'' not in sentence and one of the pronouns is in [``Sie'',``Ihr'', ``Ihre'', ``Ihren'', ``Ihrem'', ``Ihrer'', ``Ihres''], then is\_pronoun\_formal = $\textsc{true}$.
        \item If one of the pronouns is in [``du'',``dein'', ``deine'', ``deinen'', ``deinem'', ``deiner'', ``deines'', ``dich''], then is\_pronoun\_formal = $\textsc{true}$.
        \item If ``!'' in sentence one of the pronouns is in [``er'',``sie'', ``es'', ``ihr''], then is\_pronoun\_formal = $\textsc{true}$.
        \item is\_informal = is\_pronoun\_informal.
        \item is\_formal = is\_pronoun\_formal.
    \end{itemize}
    \item Keep samples if is\_formal != is\_informal, use `formal' label if is\_formal or `informal' label if is\_informal.
    \item For each sample, create context by keeping the preceding three to five English sentences, depending if word count is above 20.
\end{enumerate}

\subsubsection{``it'' resolution}

We provide the following list of steps to create the ``it'' resolution dataset. The steps apply to all languages:

\begin{enumerate}[noitemsep,partopsep=0pt,topsep=0pt,parsep=0pt,leftmargin=4mm]
    \item For each English sentence in the OpenSubtitle dataset, keep sentences where the word``it'' exists.
    \begin{itemize}[noitemsep,topsep=3pt]
        \item Using a dependency parser, if ``it'' is expletive\footnote{The \href{https://spacy.io/}{spaCy} dependency parser can be used to find expletive ``it''.}, skip sample. 
        \item In the parallel Spanish, French, German or Japanese sentence, if the sentence does not contain a verb and a gendered pronouns, skip sample.
        \item Keep gender label.
    \end{itemize}
    \item For each sample, create context by keeping the preceding three to five English sentences, depending if word count is above 20.
\end{enumerate}

\subsubsection{Gender Neutral Names}

We provide the following list of steps to create the gender neutral names dataset. Please note that for simplicity we used binary genders. Genders beyond female and male will be left for future work. The steps apply to all languages:

\begin{enumerate}[noitemsep,partopsep=0pt,topsep=0pt,parsep=0pt,leftmargin=4mm]
    \item Compile a list $L_{gnn}$ of gender neutral (unisex) names
    \begin{itemize}[noitemsep,topsep=3pt]
        \item Collect a list of names with gender statistic such as the percentage of people with the name who identify as female or male\footnote{Names with gender statistics were compiled and combined using a Japanese names database \cite{baby_ja_names_2020} and  a  \href{https://github.com/fivethirtyeight/data/tree/master/unisex-names}{English names database} that originates from the United States Social Security Administration.}.
        \item Keep the names that are used in approximately equal proportions (unisex) with at least a female or male proportion above 40\%.
    \end{itemize}
    \item For each gender neutral name $\in L_{gnn}$, find a sentence that contains the name in the English sentence and keep the corresponding parallel sentence in Spanish, French, German or Japanese.
    \begin{itemize}[noitemsep,topsep=3pt]
        \item If the English sentence has gendered pronouns, skip the sentence if multiple genders are detected.
        \item If the English sentence has no gendered pronouns, use a Part-of-Speech tagger\footnote{Language specific spaCy models could be used.} on the corresponding parallel sentence in Spanish, French, German or Japanese and skip the sentence if multiple genders are detected.
        \item Keep gender label.
    \end{itemize}
    \item Replace gendered pronouns with [pr] in the source English sentence to remove simple clues about the name's gender.
    \item For each sample, create context by keeping the succeeding three to five English sentences, depending if word count is above 20.
\end{enumerate}

\section{Prompt templates used in experiments}
\label{app:prompts}

In this section, we discuss the main prompt templates used in experiments.
This includes \icp \emph{Translator} generalist and specialist templates to ask questions about ambiguities and exemplars to translate in French, Spanish, German or Japanese.
It also includes \icp \emph{User} generalist and specialist templates to answer questions given a context.
We also provide the prompt templates for the PaLM-with-Context experiments where we use context and the same exemplars to translate in French, Spanish, German or Japanese.
Please note that we have normalized special characters for simplicity.
The German and Japanese templates as well as Spanish and French templates with special characters can be found in our public code and data repository.
In the python methods listed below, \emph{en\_text} is the input query, \emph{ctx} is the context, \emph{question} is the question from the Translator model and \emph{anwer} is the answer from the User model.

\subsection{\icp Simulated User Prompts}
\label{app:sim_template}

The $8$-shot generalist Simulated \emph{User} prompt template is the same for all languages and is provided in code block listing~\ref{lst:user_gen}.

\begin{lstlisting}[language=Python, caption=\icp Generalist Simulated User Prompt Template, label=lst:user_gen]
def generalist_simulated_user_context(en_text, question, ctx):
    """Generalist Simulated user has access to context and answers the question."""
  
    templated_input =
f"""[web] Given a Context (C), provide an Answer (A) to the Question (Q):

S: about
C: About 2% of the households are enumerated using the canvasser method.
Q: Is "about" an adverb that means approximately, near or a preposition that means regarding, over, surrounding?
A: "about" means approximately.


S: rent
C: Many single women cannot live independently because they cannot (afford to) own or rent housing
Q: Is "rent" a tenant's regular payment for a property or to pay someone for the use of something?
A: "rent" is to pay someone for the use of something.


S: abstract
C: For the international community is not an abstract concept, it consists of us ourselves.
Q: Is "abstract" to consider theoretically, to extract something, or a summary, or an adjective?
A: "abstract" is an adjective that modifies "concept" in the phrase "abstract concept".


S: What do you mean?
C: Daria, I just think that your field of vision could really be enhanced... - Come on, Mom. - It's not my field of vision you want to enhance.
Q: "you" can be neutral, formal, informal. Who does "you" refer to?
A: "you" is 'informal' since the listener is the speaker's "mom", it implies a familiarity with the listener "you".


S: This will accelerate your metabolic functions-- help you make the transition.
C: At the very least, get them to hold their fire. - Captain, the transporters are off-line. The docking port hasn't been hit yet.
Q: "you" can be neutral, formal, informal. Who does "you" refer to?
A: "you" is 'formal' since "you" refers to a Captain and the speaker will typically use a polite form.


S: You know where it begins, you never know where it ends...
C: Someone once told me we always are where we're supposed to be. - Now I believe it. - Life is a journey.
Q: "you" can be neutral, formal, informal. Who does "you" refer to in (S)?
A: "you" is \'neutral\' because it is a generic "you" that refers to people in general on their journey through life.


S: it is also very pretty.
C: Even when it is pouring outside, this umbrella is both practical and elegant.
Q: What does "it" refer to?
A: "it" is a harp.


S: Tell me, why do they have to tilt it?
C: -Frog is wrong. - I see here that you play the harp.
Q: What does "it" refer to?
A: "it" is an umbrella.


S: {en_text.strip()}
C: {ctx.strip()}
Q: {question}
A:"""
    return templated_input
\end{lstlisting}

The $8$-shot \emph{formality} specialist Simulated \emph{User} prompt template is the same for all languages and is provided in code block listing~\ref{lst:user_spe_form}.

\begin{lstlisting}[language=Python, caption=\icp \textbf{Formality} Specialist Simulated User Prompt Template, label=lst:user_spe_form]
def formality_simulated_user_context(en_text, question, ctx):
    """Formality simulated user has access to context and answers the question."""
  
    templated_input =
f"""[web] Given a Context (C), provide an Answer (A) to the Question (Q) about Sentence (S):

S: This is for you, too.
C: I'm Freya. - Welcome to Denmark, Mr. Helm. - You always greet people like this? - I'm Freya Carlson, your Tourist Bureau contact.
Q: "you" can be neutral, formal, informal. Who does "you" refer to in (S)?
A: "you" is \'formal\' since "you" refers to a customer or tourist that Freya Carlson is greeting with the polite form "Mr.".


S: - i can gladly help you.
C: I will go to town to fetch the materials. Once I return, we can repair your majesty's royal carriage.
Q: "you" can be formal or informal. Who does "you" refer to?
A: "you" is \'formal\' since "you" refers to "your majesty".


S: You know what I mean.
C: Elizabeth, will you bring the binoculars? - [Elizabeth] Mm, the stench is horrible. [John] Here, take a hold of this. - [Elizabeth] Is it dead?
Q: "you" can be neutral, formal, informal. Who does "you" refer to in (S)?
A: "you" is \'informal\' since the listener "John" has familiarity with the speaker and uses the first name "Elizabeth".


S: You think you can make it through that kind of stuff, you think you can make it through anything.
C: Well, transitions are hard. - Been together ever since college. - Been through a lot. - You know, us coming out to her family, and her brother dying.
Q: "you" can be neutral, formal, informal. Who does "you" refer to in (S)?
A: "you" is \'neutral\' because it is a generic "you" that refers to people in general going through a difficult moment.


S: You can imagine the princess-sized tantrum that followed.
Q: "you" can be neutral, formal, informal. Who does "you" refer to in (S)?
C: This is the bike that I learned to ride on. - I just didn't know my mom kept it. - It used to have these training wheels on the back with lights that would flash every time you pedaled. - Then one day, my mom took them off and said it was time to be a big girl.
A: "you" is \'informal\' since the speaker is talking about a funny childhood memory which implies a familiarity with the listener "you".


S: Can I just say, it's been an absolute pleasure to finally meet you?
C: Generations of Daleks just woke up very cross, and they're coming up the pipes. - Or to put it another way... bye! - Doctor, you must help me.
Q: "you" can be neutral, formal, informal. Who does "you" refer to in (S)?
A: "you" is \'formal\' since "you" refers to a "Doctor" that the speaker just met.


S: You know where it begins, you never know where it ends...
C: Someone once told me we always are where we're supposed to be. - Now I believe it. - Life is a journey.
Q: "you" can be neutral, formal, informal. Who does "you" refer to in (S)?
A: "you" is \'neutral\' because it is a generic "you" that refers to people in general on their journey through life.


S: City policemen questioned many of you this week.
C: Lying on his belly, he was carried home on a makeshift stretcher. - Next Sunday, after the service, the Baron asked the pastor to let him speak.
Q: "you" can be neutral, formal, informal. Who does \"you\" refer to in (S)?
A: "you" is \'formal\' since the speaker directly addresses several people or "many of you", the plural form of "you".


S: {en_text.strip()}
C: {ctx.strip()}
Q: {question}
A: """
    return templated_input
\end{lstlisting}

The $8$-shot \emph{polysemy} specialist Simulated \emph{User} prompt template is the same for all languages and is provided in code block listing~\ref{lst:user_spe_poly}.

\begin{lstlisting}[language=Python, caption=\icp \textbf{Polysemy} Specialist Simulated User Prompt Template, label=lst:user_spe_poly]
def polysemy_simulated_user_context(en_text, question, ctx):
    """Polysemy simulated user has access to context and answers the question."""
  
    templated_input =
f"""[web] Given a Context (C), provide an Answer (A) to the Question (Q):

S: abstract
C: For the international community is not an abstract concept, it consists of us ourselves.
Q: Is "abstract" to consider theoretically, to extract something, or a summary, or an adjective?
A: "abstract" is an adjective that modifies the word "concept".


S: abstract
C: We need to abstract the data from various studies.
Q: Is "abstract" to consider theoretically, to extract something, or a summary, or an adjective?
A: "abstract" means to extract something.


S: about
C: About 2% of the households are enumerated using the canvasser method.
Q: Is "about" an adverb that means approximately, near or a preposition that means regarding, over, surrounding?
A: "about" means approximately.


S: about
C: The story is about soldier returning home after the war.
Q: Is "about" an adverb that means approximately, near or a preposition that means regarding, over, surrounding?
A: "about" means regarding.


S: bank
C: The online banking application does not work. I tried a few times and I could not transfer the funds. I went to the bank.
Q: Is "bank" a financial institution, the edge of a river, a set or series of similar things or the cushion of a pool?
A: "bank" is a financial institution.


S: rent
C: Many single women cannot live independently because they cannot (afford to) own or rent housing
Q: Is "rent" a tenant's regular payment for a property or to pay someone for the use of something?
A: "rent" is to pay someone for the use of something.


S: bat
C: The bat flew over the forest and back to its cave.
Q: Is "bat" an animal or a sports equipment?
A: "bat" is an animal.


C: {ctx}
Q: {question}
A: """
    return templated_input
\end{lstlisting}

\subsection{\icp Generalist Prompt Templates for each target language}
\label{app:gen_templates}

The $8$-shot \emph{Spanish} generalist \emph{Translator} prompt template is the same for all test ambiguity data and is provided in code block listing~\ref{lst:trsl_gen_es}.

\begin{lstlisting}[language=Python, caption=\icp \textbf{Spanish} Generalist Translator Prompt Template, label=lst:trsl_gen_es]
def spanish_generalist_translator_interactive(en_text, question=None, answer=None):
    """Translation model asks questions and uses answers to translate"""
    if answer == None:
        #  Ask questions
        instructions = "[web] Given sentence 'S' to translate to Spanish, ask clarifying questions 'Q' to clarify ambiguities or multiple senses:"
    else:
        #  Translate given answer
        instructions = "[web] Given answer 'U' to question 'Q', provide the Spanish translation 'A' of sentence 'S'. Provide the best answer:"
 
    templated_input =
"""

S: about
Q: Is "about" an adverb that means approximately, near or a preposition that means regarding, over, surrounding?%s


S: rent
Q: Is "rent" a tenant's regular payment for a property or to pay someone for the use of something?%s


S: abstract
Q: Is "abstract" to consider theoretically, to extract something, or a summary, or an adjective?%s


S: You think if I get contacts I'll suddenly turn into the homecoming queen.
Q: "you" can be neutral, formal, informal. Who does "you" refer to?%s


S: This will accelerate your metabolic functions-- help you make the transition.
Q: "you" can be neutral, formal, informal. Who does "you" refer to?%s


S: They could wait 'till you're on the beach, then cut loose, or start firing right away.
Q: "you" can be neutral, formal, informal. Who does "you" refer to?%s


S: can't they just build it on an angle?
Q: What does "it" refer to?%s


S: It is also very pretty.
Q: What does "it" refer to?%s


"""
    if answer is None:
        templated_input = templated_input % ('', '', '', '', '', '', '', '')
        templated_input = f"{instructions}\n" + templated_input + f"S: {en_text}\nQ:"
    else:
        templated_input = templated_input % (
            '\nU: "about" means approximately.\nA: aproximadamente, cerca de, alrededor de, casi, mas o menos',
            '\nU: "rent" is to pay someone for the use of something.\nA: alquilar, arrendar, rentar',
            '\nU: "abstract" is an adjective that modifies "concept" in the phrase "abstract concept".\nA: abstraccion, abstracto',
            '\nU: "you" is \'informal\' since the listener is the speaker\'s "mom", it implies a familiarity with the listener "you".\nA: Tu piensas que si uso lentes de contacto de repente me convertire en la nueva reina del colegio.',
            '\nU: "you" is \'formal\' since "you" refers to a Captain and the speaker will typically use a polite form.\nA: Esto acelerara sus funciones metabolicas. Lo ayudara a hacer la transicion.',
            '\nU: "you" is \'neutral\' because it is a generic "you" that refers to people in general and not someone specific.\nA: Podian aguardar a que uno estuviera en la playa y atacar o comenzar a disparar.',
            '\nU: "it" is a harp.\nA: no pueden hacerla en angulo?',
            '\nU: "it" is an umbrella.\nA: Es muy bonita tambien.',
        ) 
    templated_input = f"{instructions}\n" + templated_input + f"S: {en_text}\nQ: {question}\nU: {answer}\nA: "
    return templated_input
\end{lstlisting}

The $8$-shot \emph{French} generalist \emph{Translator} prompt template is the same for all test ambiguity data and is provided in code block listing~\ref{lst:trsl_gen_fr}.

\begin{lstlisting}[language=Python, caption=\icp \textbf{French} Generalist Translator Prompt Template, label=lst:trsl_gen_fr]
def french_generalist_translator_interactive(en_text, question=None, answer=None):
    """Translation model asks questions and uses answers to translate"""
    if answer == None:
        #  Ask questions
        instructions = "[web] Given sentence 'S' to translate to French, ask clarifying questions 'Q' to clarify ambiguities or multiple senses:"
    else:
        #  Translate given answer
        instructions = "[web] Given answer 'U' to question 'Q', provide the French translation 'A' of sentence 'S'. Provide the best answer:"

    templated_input = """

S: about
Q: Is "about" an adverb that means approximately, near or a preposition that means regarding, over, surrounding?%s


S: rent
Q: Is "rent" a tenant's regular payment for a property or to pay someone for the use of something?%s


S: abstract
Q: Is "abstract" to consider theoretically, to extract something, or a summary, or an adjective?%s


S: You know where it begins, you never know where it ends...
Q: "you" can be neutral, formal, informal. Who does "you" refer to?%s


S: This is for you, too.
Q: "you" can be neutral, formal, informal. Who does "you" refer to?%s


S: You know where it begins, you never know where it ends...
Q: "you" can be neutral, formal, informal. Who does "you" refer to?%s


S: I'll help you find it before [pr] does.
Q: What does "it" refer to?%s


S: [pr] must have forced it somehow.
Q: What does "it" refer to?%s


"""

    if answer is None:
        templated_input = templated_input % ('', '', '', '', '', '', '', '')
        templated_input = f"{instructions}\n" + templated_input + f"S: {en_text}\nQ:"
    else:
        templated_input = templated_input % (
        '\nU: "about" means approximately.\nA: environ, presque, quelque, a peu pres, approximativement',
        '\nU: "rent" is to pay someone for the use of something.\nA: louer',
        '\nU: "abstract" is an adjective that modifies "concept" in the phrase "abstract concept".\nA: abstraction, abstrait',
        '\nU: "you" is \'informal\' since the speaker has familiarity with the listener and uses the first name "Jerry".\nA: A qui as-tu parle ?',
        '\nU: "you" is \'formal\' since "you" refers to a customer or tourist that Freya Carlson is greeting with the polite form "Mr.".\nA: Ceci est pour vous.',
        '\nU: "you" is \'neutral\' because it is a generic "you" that refers to people in general going through a difficult moment.\nA: On sait ou cela commence, mais on ne sait jamais ou cela se termine...',
        '\nU: "it" is a key.\nA: Je vous aiderai a la trouver avant elle.',
        '\nU: "it" is a gate.\nA: Il a du le forcer d\'une maniere ou d\'une autre.',
        ) 
    templated_input = f"{instructions}\n" + templated_input + f"S: {en_text}\nQ: {question}\nU: {answer}\nA: "
    return templated_input
\end{lstlisting}

\subsection{\icp Specialist Prompt Templates for each target language}
\label{app:spe_templates}

The \emph{Spanish formality} specialist \emph{Translator} prompt template is the same for all test ambiguity data and is provided in code block listing~\ref{lst:trsl_spe_form_es}.

\begin{lstlisting}[language=Python, caption=\icp \textbf{Spanish Formality} Specialist Translator Prompt Template, label=lst:trsl_spe_form_es]
def spanish_formality_translator_interactive(en_text, question=None, answer=None):
    """Translation model asks questions and uses answers to translate"""
    if answer == None:
        #  Ask questions
        instructions = "[web] Given sentence 'S' to translate to Spanish, ask clarifying questions 'Q' to clarify ambiguities or multiple senses:"
    else:
        #  Translate given answer
        instructions = "[web] Given answer 'U' to question 'Q', provide the Spanish translation 'A' of sentence 'S'. Provide the best answer:"

    templated_input = """

S: This will accelerate your metabolic functions-- help you make the transition.
Q: "you" can be neutral, formal, informal. Who does "you" refer to?%s


S: Poor baby... here's yours!
Q: "you" can be neutral, formal, informal. Who does "you" refer to?%s


S: They could wait 'till you're on the beach, then cut loose, or start firing right away.
Q: "you" can be neutral, formal, informal. Who does "you" refer to?%s


S: You think if I get contacts I'll suddenly turn into the homecoming queen.
Q: "you" can be neutral, formal, informal. Who does "you" refer to?%s


S: For centuries, we have watched you, listened to your radio signals and learned your speech and your culture.
Q: "you" can be neutral, formal, informal. Who does "you" refer to?%s


S: I never have. I'm not sure you're supposed to.
Q: "you" can be neutral, formal, informal. Who does "you" refer to?%s


"""

    if answer is None:
        templated_input = templated_input % ('', '', '', '', '', '')
        templated_input = f"{instructions}\n" + templated_input + f"S: {en_text}\nQ:"
    else:
        templated_input = templated_input % (
        '\nU: "you" is \'formal\' since "you" refers to a Captain and the speaker will typically use a polite form.\nA: Esto acelerara sus funciones metabolicas. Lo ayudara a hacer la transicion.',
        '\nU: "you" is \'informal\' since the speaker has familiarity with the listener and they both use "baby" and "buddy" to address each other.\nA: Pobre bebe... aqui esta el tuyo!',
        '\nU: "you" is \'neutral\' because it is a generic "you" that refers to people in general and not someone specific.\nA: Podian aguardar a que uno estuviera en la playa y atacar o comenzar a disparar.',
        '\nU: "you" is \'informal\' since the listener is the speaker\'s "mom", it implies a familiarity with the listener "you".\nA: Tu piensas que si uso lentes de contacto de repente me convertire en la nueva reina del colegio.',
        '\nU: "you" is \'formal\' since the speaker addresses people not acquainted with or unfamiliar.\nA: Durante siglos, los hemos observado, escuchado sus senales de radio. Hemos aprendido su idioma y cultura.',
        '\nU: "you" is \'neutral\' because it is a generic "you" that refers to people in general that have been in this "line of work".\nA: Yo no. No creo que uno deba acostumbrarse.'
        ) 
    templated_input = f"{instructions}\n" + templated_input + f"S: {en_text}\nQ: {question}\nU: {answer}\nA: "
    return templated_input
\end{lstlisting}

The \emph{Spanish polysemy} specialist \emph{Translator} prompt template is the same for all test ambiguity data and is provided in code block listing~\ref{lst:trsl_spe_poly_es}. Please note that the instructions for the translation step is different than the generalist or the formality specialist template.

\begin{lstlisting}[language=Python, caption=\icp \textbf{Spanish Polysemy} Specialist Translator Prompt Template, label=lst:trsl_spe_poly_es]
def spanish_polysemy_translator_interactive(en_text, question=None, answer=None):
    """Translation model asks questions and uses answers to translate"""
    if answer == None:
        #  Ask questions
        instructions = "[web] Given an English word 'S' to translate to Spanish, to clarify ambiguities and understand multiple senses ask questions 'Q':"
    else:
        #  Translate given answer
        instructions = "[web] Given answer 'U' to question 'Q', Translate word 'S' into Spanish and provide unique and non-repeating synonyms in 'A':"

    templated_input = """

S: abstract
Q: Is "abstract" to consider theoretically, to extract something, or a summary, or an adjective?%s


S: abstract
Q: Is "abstract" to consider theoretically, to extract something, or a summary, or an adjective?%s


S: about
Q: Is "about" an adverb that means approximately, near or a preposition that means regarding, over, surrounding?%s


S: bank
Q: Is "bank" to tilt sideways, or a financial institution, the edge of a river, a set or series of similar things or the cushion of a pool?%s


S: rent
Q: Is "rent" a tenant's regular payment for a property or to pay someone for the use of something?%s


"""

    if answer is None:
        templated_input = templated_input % ('', '', '', '', '')
        templated_input = f"{instructions}\n" + templated_input + f"S: {en_text}\nQ: "
    else:
        templated_input = templated_input % (
        '\nU: "abstract" is an adjective that modifies "concept" in the phrase "abstract concept".\nA: abstraccion, abstracto',
        '\nU: "abstract" means to extract something.\nA: abstraer',
        '\nU: "about" means approximately.\nA: aproximadamente, cerca de, alrededor de, casi, mas o menos',
        '\nU: "bank" is a financial institution.\nA: banco',
        '\nU: "rent" is to pay someone for the use of something.\nA: alquilar, arrendar, rentar'
        )
    templated_input = f"{instructions}\n" + templated_input + f"S: {en_text}\nQ: {question}\nU: {answer}\nA: "
    return templated_input
\end{lstlisting}

The \emph{French formality} specialist \emph{Translator} prompt template is the same for all test ambiguity data and is provided in code block listing~\ref{lst:trsl_spe_form_fr}.

\begin{lstlisting}[language=Python, caption=\icp \textbf{French Formality} Specialist Translator Prompt Template, label=lst:trsl_spe_form_fr]
def french_formality_translator_interactive(en_text, question=None, answer=None):
    """Translation model asks questions and uses answers to translate"""
    if answer == None:
        #  Ask questions
        instructions = "[web] Given sentence 'S' to translate to French, ask clarifying questions 'Q' to clarify ambiguities or multiple senses:"
    else:
        #  Translate given answer
        instructions = "[web] Given answer 'U' to question 'Q', provide the French translation 'A' of sentence 'S'. Provide the best answer:"

    templated_input = """

S: This is for you, too.
Q: "you" can be neutral, formal, informal. Who does "you" refer to?%s


S: To whom have you been talking?
Q: "you" can be neutral, formal, informal. Who does "you" refer to?%s


S: You know where it begins, you never know where it ends...
Q: "you" can be neutral, formal, informal. Who does "you" refer to?%s


S: You can imagine the princess-sized tantrum that followed.
Q: "you" can be neutral, formal, informal. Who does "you" refer to?%s


S: City policemen questioned many of you this week.
Q: "you" can be neutral, formal, informal. Who does "you" refer to?%s


S: You think you can make it through that kind of stuff, you think you can make it through anything.
Q: "you" can be neutral, formal, informal. Who does "you" refer to?%s


"""

    if answer is None:
        templated_input = templated_input % ('', '', '', '', '', '')
        templated_input = f"{instructions}\n" + templated_input + f"S: {en_text}\nQ:"
    else:
        templated_input = templated_input % (
        '\nU: \nA: Ceci est pour vous.',
        '\nU: \nA: A qui as-tu parle ?',
        '\nU: \nA: On sait ou cela commence, mais on ne sait jamais ou cela se termine...',
        '\nU: \nA: Tu peux imaginer la colere de princesse qui a suivi.',
        '\nU: \nA: Les gendarmes sont venus interroger nombre d\'entre vous.',
        '\nU: \nA: On pense que quand on arrive a traverser ce genre de chose, on peut traverser n\'importe quoi.'
        )
    templated_input = f"{instructions}\n" + templated_input + f"S: {en_text}\nQ: {question}\nU: {answer}\nA: "
    return templated_input
\end{lstlisting}

The \emph{French polysemy} specialist \emph{Translator} prompt template is the same for all test ambiguity data and is provided in code block listing~\ref{lst:trsl_spe_poly_fr}. Please note that the instructions for the translation step is different than the generalist or the formality specialist template.

\begin{lstlisting}[language=Python, caption=\icp \textbf{French Polysemy} Specialist Translator Prompt Template, label=lst:trsl_spe_poly_fr]
def french_polysemy_translator_interactive(en_text, question=None, answer=None):
    """Translation model asks questions and uses answers to translate"""
    if answer == None:
        #  Ask questions
        instructions = "[web] Given an English word 'S' to translate to French, to clarify ambiguities and understand multiple senses ask questions 'Q':"
    else:
        #  Translate given answer
        instructions = "[web] Given answer 'U' to question 'Q', Translate word 'S' into French and provide unique and non-repeating synonyms in 'A':"

    templated_input = """

S: abstract
Q: Is "abstract" to consider theoretically, to extract something, or a summary, or an adjective?%s


S: abstract
Q: Is "abstract" to consider theoretically, to extract something, or a summary, or an adjective?%s


S: about
Q: Is "about" an adverb that means approximately, near or a preposition that means regarding, over, surrounding?%s


S: bank
Q: Is "bank" to tilt sideways, or a financial institution, the edge of a river, a set or series of similar things or the cushion of a pool?%s


S: rent
Q: Is "rent" a tenant's regular payment for a property or to pay someone for the use of something?%s


"""

    if answer is None:
        templated_input = templated_input % ('', '', '', '', '')
        templated_input = f"{instructions}\n" + templated_input + f"S: {en_text}\nQ: "
    else:
        templated_input = templated_input % (
        '\nU: "abstract" is an adjective that modifies "concept" in the phrase "abstract concept".\nA: abstraction, abstrait', 
        '\nU: "abstract" means to extract something.\nA: abstraire, extraire',
        '\nU: "about" means approximately.\nA: environ, presque, quelque, a peu pres, approximativement',
        '\nU: "bank" is a financial institution.\nA: banque',
        '\nU: "rent" is to pay someone for the use of something.\nA: louer'
        )
    templated_input = f"{instructions}\n" + templated_input + f"S: {en_text}\nQ: {question}\nU: {answer}\nA: "
    return templated_input
\end{lstlisting}

\subsection{PaLM-with-Context Generalist Prompt Templates for each target language}
\label{app:gen_templates_baseline}

The $8$-shot PaLM-with Context \emph{Spanish} generalist prompt template is the same for all test ambiguity data and is provided in code block listing~\ref{lst:palm_gen_es}.

\begin{lstlisting}[language=Python, caption=PaLM-with-Context \textbf{Spanish} Generalist Prompt Template, label=lst:palm_gen_es]
def spanish_baseline_generalist_translator_context(en_text, ctx):
    """Translation model uses context to translate."""

    templated_input = f"""[web] Given context 'C', Translate 'T' from English to Spanish:

C: About 2% of the households are enumerated using the canvasser method.
T: about
A: aproximadamente, cerca de, alrededor de, casi, mas o menos


C: Many single women cannot live independently because they cannot (afford to) own or rent housing
T: rent
A: alquilar, arrendar, rentar


C: For the international community is not an abstract concept, it consists of us ourselves.
T: abstract
A: abstraccion, abstracto


C: Daria, I just think that your field of vision could really be enhanced... - Come on, Mom. - It's not my field of vision you want to enhance. - What do you mean?
T: You think if I get contacts I'll suddenly turn into the homecoming queen.
A: Tu piensas que si uso lentes de contacto de repente me convertire en la nueva reina del colegio.


C: At the very least, get them to hold their fire. - Captain, the transporters are off-line. - The docking port hasn't been hit yet.
T: This will accelerate your metabolic functions-- help you make the transition.
A: Esto acelerara sus funciones metabolicas. Lo ayudara a hacer la transicion


C: Some of the guys got a little sick. - They were scared; I was scared. - I don't think we had any reason to be otherwise.
T: They could wait 'till you're on the beach, then cut loose, or start firing right away.
A: Podian aguardar a que uno estuviera en la playa y atacar o comenzar a disparar.


C: Even when it is pouring outside, this umbrella is both practical and elegant.
T: It is also very pretty.
A: Es muy bonita tambien.


C: -Frog is wrong. - I see here that you play the harp. - Tell me, why do they have to tilt it?
T: can't they just build it on an angle?
A: no pueden hacerla en angulo?


C: {ctx}
T: {en_text}
A:"""
    return templated_input
\end{lstlisting}

The $8$-shot PaLM-with Context \emph{French} generalist prompt template is the same for all test ambiguity data and is provided in code block listing~\ref{lst:palm_gen_fr}.

\begin{lstlisting}[language=Python, caption=PaLM-with-Context \textbf{French} Generalist Prompt Template, label=lst:palm_gen_fr]
def french_baseline_generalist_translator_context(en_text, ctx):
    """Translation model uses context to translate."""

    templated_input = f"""[web] Given context 'C', Translate 'T' from English to French:

C: About 2% of the households are enumerated using the canvasser method.
T: about
A: environ, presque, quelque, a peu pres, approximativement


C: Many single women cannot live independently because they cannot (afford to) own or rent housing
T: rent
A: louer


C: For the international community is not an abstract concept, it consists of us ourselves.
T: abstract
A: abstraction, abstrait


C: I believe! - -Who else knows? - -I don't know. - Jerry, names! - I don't want to dance!
T: To whom have you been talking?
A: A qui as-tu parle ?


C: I'm Freya. - Welcome to Denmark, Mr. Helm. - You always greet people like this? - I'm Freya Carlson, your Tourist Bureau contact. - These are for you. Street maps, places of interest.
T: This is for you, too.
A: Ceci est pour vous.


C: It's like the city's changed her. - Well, transitions are hard. - Been together ever since college. - Been through a lot. - You know, us coming out to her family, and her brother dying.
T: You know where it begins, you never know where it ends...
A: On sait ou cela commence, mais on ne sait jamais ou cela se termine...


C: Even when it is pouring outside, this umbrella is both practical and elegant.
T: it is also very pretty.
A: il est aussi tres beau.


C: Okay, you don't smash the cherry on that. Just plop it in at the end.
T: Try to keep it in the top of the glass.
A: Essaie de la garder dans le haut du verre.


C: {ctx}
T: {en_text}
A:"""
    return templated_input
\end{lstlisting}

\subsection{PaLM-with-Context Specialist Prompt Templates for each target language}
\label{app:spe_templates_baseline}

The PaLM-with Context \emph{Spanish Formality} specialist prompt template is the same for all test ambiguity data and is provided in code block listing~\ref{lst:palm_spe_form_es}.

\begin{lstlisting}[language=Python, caption=PaLM-with-Context \textbf{Spanish Formality} Specialist Prompt Template, label=lst:palm_spe_form_es]
def spanish_baseline_formality_translator_context(en_text, ctx):
    """Translation model uses context to translate."""

    templated_input = f"""[web] Given context 'C', Translate 'T' from English to Spanish:

C: At the very least, get them to hold their fire. - Captain, the transporters are off-line. - The docking port hasn't been hit yet.
T: This will accelerate your metabolic functions-- help you make the transition.
A: Esto acelerara sus funciones metabolicas. Lo ayudara a hacer la transicion.

C: Who? - Me! - I think I've got a cold. - "Hey buddy, give me a Magic Hug will you!" - Magic Hug! - And me? - Shut up Swami
T: Poor baby... here's yours!
A: Pobre bebe... aqui esta el tuyo!

C: Some of the guys got a little sick. - They were scared; I was scared. - I don't think we had any reason to be otherwise.
T: They could wait 'till you're on the beach, then cut loose, or start firing right away.
A: Podian aguardar a que uno estuviera en la playa y atacar o comenzar a disparar.

C: Daria, I just think that your field of vision could really be enhanced... - Come on, Mom. - It's not my field of vision you want to enhance. - What do you mean?
T: You think if I get contacts I'll suddenly turn into the homecoming queen.
A: Tu piensas que si uso lentes de contacto de repente me convertire en la nueva reina del colegio.

C: Men of earth, we of the planet Mars give you this warning. - We have known your planet earth since the first creature crawled out of the primeval slime of your seas to become man.
T: For centuries, we have watched you, listened to your radio signals and learned your speech and your culture.
A: Durante siglos, los hemos observado, escuchado sus senales de radio. Hemos aprendido su idioma y cultura.

C: Pull over here. This is the spot. - I guess you run into a lot of dead bodies in your line of work. - You get used to it.
T: I never have. I'm not sure you're supposed to.
A: Yo no. No creo que uno deba acostumbrarse.

C: {ctx}
T: {en_text}
A:"""
    return templated_input
\end{lstlisting}

The PaLM-with Context \emph{Spanish Polysemy} specialist prompt template is the same for all test ambiguity data and is provided in code block listing~\ref{lst:palm_spe_poly_es}.

\begin{lstlisting}[language=Python, caption=PaLM-with-Context \textbf{Spanish Polysemy} Specialist Prompt Template, label=lst:palm_spe_poly_es]
def spanish_baseline_polysemy_translator_context(en_text, ctx):
    """Translation model uses context to translate."""

    templated_input = f"""[web] Given context 'C', Translate 'T' from English to Spanish:


C: Many single women cannot live independently because they cannot (afford to) own or rent housing
T: rent
A: alquilar, arrendar, rentar


C: We need to abstract the data from various studies.
T: abstract
A: abstraer


C: About 2% of the households are enumerated using the canvasser method.
T: about
A: aproximadamente, cerca de, alrededor de, casi, mas o menos


C: The bat flew over the forest and back to its cave.
T: bat
A: murcielago


C: For the international community is not an abstract concept, it consists of us ourselves.
T: abstract
A: abstraccion, abstracto


C: {ctx}
T: {en_text}
A:"""
    return templated_input
\end{lstlisting}

The PaLM-with Context \emph{French Formality} specialist prompt template is the same for all test ambiguity data and is provided in code block listing~\ref{lst:palm_spe_form_fr}.

\begin{lstlisting}[language=Python, caption=PaLM-with-Context \textbf{French Formality} Specialist Prompt Template, label=lst:palm_spe_form_fr]
def french_baseline_formality_translator_context(en_text, ctx):
    """Translation model uses context to translate."""

    templated_input = f"""[web] Given context 'C', Translate 'T' from English to French:

C: I'm Freya. - Welcome to Denmark, Mr. Helm. - You always greet people like this? - I'm Freya Carlson, your Tourist Bureau contact. - These are for you. Street maps, places of interest.
T: This is for you, too.
A: Ceci est pour vous.

C: I believe! - -Who else knows? - -I don't know. - Jerry, names! - I don't want to dance!
T: To whom have you been talking?
A: A qui as-tu parle ?

C: It's like the city's changed her. - Well, transitions are hard. - Been together ever since college. - Been through a lot. - You know, us coming out to her family, and her brother dying.
T: You know where it begins, you never know where it ends...
A: On sait ou cela commence, mais on ne sait jamais ou cela se termine...

C: You know, if you're gonna go for a spin, I suggest you get your helmet. - This is the bike that I learned to ride on. - I just didn't know my mom kept it. - It used to have these training wheels on the back with lights that would flash every time you pedaled. - Then one day, my mom took them off and said it was time to be a big girl.
T: You can imagine the princess-sized tantrum that followed.
A: Tu peux imaginer la colere de princesse qui a suivi.

C: He was in a state of shock, unable to walk. - Lying on his belly, he was carried home on a makeshift stretcher. - Next Sunday, after the service, the Baron asked the pastor to let him speak.
T: City policemen questioned many of you this week.
A: Les gendarmes sont venus interroger nombre d\'entre vous.

C: I tried to explain... He might have gotten hurt! - I was actually doing him a favour. - Someone once told me we always are where we're supposed to be. - Now I believe it. - Life is a journey.
T: You think you can make it through that kind of stuff, you think you can make it through anything.
A: On pense que quand on arrive a traverser ce genre de chose, on peut traverser n\'importe quoi.

C: {ctx}
T: {en_text}
A:"""
    return templated_input
\end{lstlisting}

The PaLM-with Context \emph{French Polysemy} specialist prompt template is the same for all test ambiguity data and is provided in code block listing~\ref{lst:palm_spe_poly_fr}.

\begin{lstlisting}[language=Python, caption=PaLM-with-Context \textbf{French Polysemy} Specialist Prompt Template, label=lst:palm_spe_poly_fr]
def french_baseline_polysemy_translator_context(en_text, ctx):
    """Translation model uses context to translate."""

    templated_input = f"""[web] Given context 'C', Translate 'T' from English to French:

C: Consequently a strategy has been defined that allows departments to approach its implementation in a step-wise manner.
T: approach
A: s'approcher, aborder, contacter, s'adresser

C: We need to abstract the data from various studies.
T: abstract
A: abstraire, extraire

C: About 2% of the households are enumerated using the canvasser method.
T: about
A: environ, presque, quelque, a peu pres, approximativement

C: The bat flew over the forest and back to its cave.
T: bat
A: chauve-souris

C: For the international community is not an abstract concept, it consists of us ourselves.
T: abstract
A: abstraction, abstrait

C: {ctx}
T: {en_text}
A:"""
    return templated_input
\end{lstlisting}

\section{More details on gender and formality classifier}
\label{app:classifier}

The classifiers fall into 2 categories: 
(1) heuristic based classification, that use the same language rules from section \ref{app:amb_heuristics};
(2) neural network based classification, using a PaLM 62B model with $8$-shot in-demonstration exemplars.
We provide below the exemplars that were used to classify gender of French in code block listing~\ref{lst:palm_gender_fr} and Spanish sentences in code block listing~\ref{lst:palm_gender_es}.
Note that we added exemplars until we had a satisfactory score on our ground truth translated sentence (see Table~\ref{tab:gender_classification}).

\begin{lstlisting}[language=Python, caption=PaLM prompt template for gender classification of French sentences, label=lst:palm_gender_fr]
def french_gender_it_classifier_template(en_text, fr_text):
  """Classify a French sentence as feminine or masculine. 7-shot examples"""

    templated_input = 
f"""[web] Given French sentence 'F', provide the gender of "it" in English sentence 'T' and explain in 'E'. Gender in 'A' must be 'feminine', 'masculine' or 'neutral':


T: lily and marshall decided to sell it for one simple reason.
F: lyly et marshall l\'avaient mise en vente pour une seule raison.
A: feminine
E: It is 'feminine' since "mise" refers to a feminine object.


T: - maybe you need to shake it up.
F: - peut-etre qu'il faut le secouer.
A: masculine
E: It is 'masculine' since "le" refers to a masculine object.


T: i want you to get it for me.
F: Je veux que tu me la rapportes.
A: feminine
E: It is 'feminine' since "la" refers to a feminine object.


T: put it back.
F: repose-le.
A: masculine
E: It is 'masculine' since "le" refers to a masculine object.


T: I'm afraid i won't be able to get it for you.
F: Je crains de ne pas pouvoir te l'obtenir.
A: neutral
E: It is 'neutral' since we cannot determine gender with "l\'" only.


T: that view is even more beautiful when you have someone to share it with.
F: elle est encore plus belle si on n'est pas seul.
A: feminine
E: It is 'feminine' since "it" refers to "view" in English and "vue" in French which is feminine.


T: where's it going?
F: ou va-t-il ?
A: masculine
E: It is 'masculine' since "it" refers to "il" in French which is masculine.


T: {en_text}
F: {fr_text}
A: """
    return templated_input
\end{lstlisting}

\begin{lstlisting}[language=Python, caption=PaLM prompt template for gender classification of Spanish sentences, label=lst:palm_gender_es]
def spanish_gender_it_classifier_template(en_text, es_text):
    """Classify a Spanish sentence as feminine or masculine. 8-shot examples"""

    templated_input = 
  
f"""[web] Given Spanish sentence 'F', provide the gender in 'A' and explain in 'E'. Gender 'A' must be either 'feminine' or 'masculine':

F: nos habriamos pasado el dia mirandola.
A: feminine
E: It is 'feminine' since "la" and verb "mirandola" refer to a feminine object.


F: - los peruanos no podian pronunciarlo.
A: masculine
E: It is 'masculine' since "lo" in verb "pronunciarlo" refers to a masculine object.


F: Quiero decir, me encantaria volver a verlo.
A: masculine
E: It is 'masculine' since "lo" in verb "verlo" refers to a masculine object.


F: debemos ponerla de vuelta?
A: feminine
E: It is 'feminine' since "la" in verb "ponerla" refers to a feminine object.


F: -tiene que bebersela o tirarla.
A: feminine
E: It is 'feminine' since "la" in verbs "bebersela" and "tirarla" refer to a feminine object.


F: Guardalo para el proximo barco.
A: masculine
E: It is 'masculine' since "lo" in verb "Guardalo" refers to a masculine object.


F: \"escuchandola me dan ganas de vivir.\"
A: feminine
E: It is 'feminine' since "la" in verb "escuchandola" refers to a feminine object.


F: !cambialo al menos!
A: masculine
E: It is 'masculine' since "lo" in verb "cambialo" refers to a masculine object.


F: {es_text.lower()}
A: """
    return templated_input
\end{lstlisting}

We have added the classification heuristics and other classification templates to our public data and code repository.
\begin{wraptable}{R}{6cm}
\caption{
PaLM 62B gender classification results on a 100 generated translation samples.
} \label{tab:gender_classification}
\begin{tabular}{ c | c}
\toprule
\textbf{Spanish} &  \textbf{French} \\
\midrule
97\% & 93\% \\
\bottomrule
\end{tabular}
\end{wraptable}

\end{document}